\documentclass[runningheads]{llncs}

\usepackage{eccv}

\usepackage{eccvabbrv}

\usepackage{graphicx}
\usepackage{booktabs}
\usepackage{amsmath}
\usepackage{amssymb}
\usepackage{multirow}
\usepackage{microtype}
\usepackage{overpic}
\usepackage{mathtools}
\usepackage{stackengine}
\usepackage{scalerel}
\usepackage[table]{xcolor}
\usepackage{colortbl}
\usepackage{wrapfig}

\usepackage{bm}

\renewcommand{\vec}[1]{\mathbf{#1}}
\newcommand{\mat}[1]{\mathbf{#1}}
\newcommand{\set}[1]{\mathcal{#1}}

\newcommand{\Hum}{H}

\newcommand{\Scn}{S}

\newcommand{\gmean}{\bm{\mu}}

\newcommand{\gskin}{\vec{w}}

\newcommand{\rasterize}{\set{R}}
\newcommand{\deform}{\set{T}}

\newcommand{\pose}{\bm{\theta}}
\newcommand{\shape}{\bm{\beta}}

\newcommand{\camerapose}{\bm{\gamma}}

\newcommand{\Normal}{\mathcal{N}}
\newcommand{\Loss}{\mathcal{L}}

\newcommand{\func}[1]{\mathsf{#1}}

\newcommand{\expr}{\bm{\psi}}

\newcommand{\had}{\odot}
\newcommand{\dd}{\mathrm{d}}

\usepackage[breaklinks,colorlinks,citecolor=eccvblue]{hyperref}

\usepackage{orcidlink}

\usepackage[capitalize,noabbrev]{cleveref}

\newcommand\blfootnote[1]{%
  \begingroup
  \renewcommand\thefootnote{}\footnote{#1}%
  \addtocounter{footnote}{-1}%
  \endgroup
}

\makeatletter
\let\@oldmaketitle\@maketitle
\renewcommand{\@maketitle}{
	\@oldmaketitle
	\begin{center}
        \includegraphics[width=\linewidth]{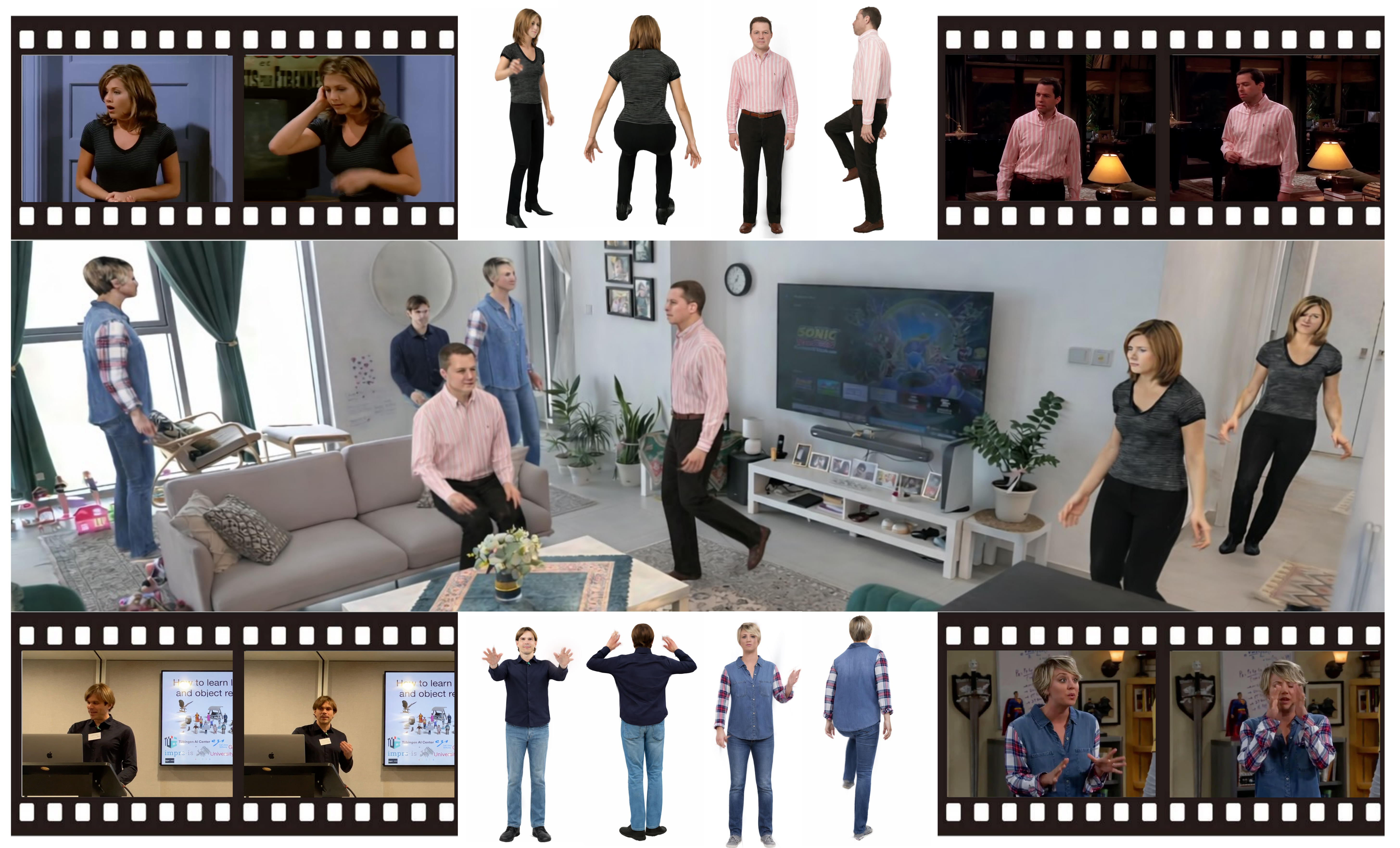}
	\end{center}
    \vspace{-6mm}
    \captionof{figure}{Given a monocular YouTube video with occlusion, AHOY reconstructs a complete, animatable 3D human avatar using video diffusion priors and 3D Gaussian Splatting, enabling photorealistic human animation within 3D scenes.}
\label{fig:teaser}

\vspace{-4mm}
}

\begin{document}

\title{AHOY! \texorpdfstring{\underline{A}}{A}nimatable \texorpdfstring{\underline{H}}{H}umans under \texorpdfstring{\underline{O}}{O}cclusion from \texorpdfstring{\underline{Y}}{Y}ouTube Videos with Gaussian Splatting and Video Diffusion Priors}

\titlerunning{AHOY! Animatable Humans under Occlusion from YouTube Videos}

\author{Aymen Mir\inst{1} \and
Riza Alp Guler\inst{3} \and
Xiangjun Tang\inst{4} \and \newline
Peter Wonka\inst{4}$^{\dagger}$ \and
Gerard Pons-Moll\inst{1,2}$^{\dagger}$}

\authorrunning{A.~Mir et al.}

\institute{T\"ubingen AI Center, University of T\"ubingen, Germany \and
Max Planck Institute for Informatics, Saarland Informatics Campus, Germany \and
Imperial College London, UK \and
King Abdullah University of Science and Technology (KAUST), Saudi Arabia}

\maketitle
\blfootnote{$^{\dagger}$ Equal advising.}

\begin{abstract}
We present AHOY, a method for reconstructing complete, animatable 3D Gaussian avatars from in-the-wild monocular video despite heavy occlusion. Existing methods assume unoccluded input: a fully visible subject, often in a canonical pose. This excludes the vast majority of real-world footage where people are routinely occluded by furniture, objects, or other people. Reconstructing from such footage poses fundamental challenges: large body regions may never be observed, and multi-view supervision per pose is unavailable.
We address these challenges with four contributions: (i)~a hallucination-as-supervision pipeline that uses identity-finetuned diffusion models to generate dense supervision for previously unobserved body regions; (ii)~a two-stage canonical-to-pose-dependent architecture that bootstraps from sparse observations to full pose-dependent Gaussian maps; (iii)~a map-pose/LBS-pose decoupling that absorbs multi-view inconsistencies from the generated data; (iv)~a head/body split supervision strategy that preserves facial identity. We evaluate on YouTube videos and on multi-view capture data with significant occlusion and demonstrate state-of-the-art reconstruction quality. We also demonstrate that the resulting avatars are robust enough to be animated with novel poses and composited into 3DGS scenes captured using cell-phone video. Our project page is available at \url{https://miraymen.github.io/ahoy/}.
\end{abstract}

\vspace{-8mm}
\section{Introduction}
\label{sec:intro}

Reconstructing photorealistic, animatable 3D humans from monocular input is a long-standing goal in computer vision and graphics, with applications spanning gaming, virtual reality, and film production. Recent methods based on NeRFs~\cite{peng2021animatable, weng_humannerf_2022_cvpr, haberman2023hdhumans} and 3DGS~\cite{kocabas2023hugs, qian20233dgsavatar, moreau2024human, zielonka2023drivable, qiu2025lhm, zhuang2025idol, huang2025adahuman, dong2025moga} have made remarkable progress, producing high-fidelity avatars that can be re-posed in real time. Yet these methods assume unoccluded monocular input, or a fully visible subject, often in a canonical pose.

These assumptions exclude the vast majority of real-world footage. YouTube alone hosts billions of videos of people in everyday activities (cooking, exercising, conversing) where the body is occluded by furniture, other people, or the scene itself. The viewpoints are limited, and large portions of the body may never be observed or cut off in the camera shot. If we could reconstruct complete, animatable avatars from such videos, we would unlock an essentially unlimited source of diverse human appearance (every body type and clothing style) without controlled capture, multi-view rigs, or cooperation from the subject.

In this work, we present \textbf{AHOY} (\textbf{A}nimatable \textbf{H}umans under \textbf{O}cclusion from \textbf{Y}ouTube Videos), a method that reconstructs complete, animatable 3D Gaussian avatars from in-the-wild monocular video despite heavy occlusion, and animates them within photorealistic 3D scenes (Fig.~\ref{fig:teaser}). Achieving this requires overcoming several challenges that do not arise when the subject is fully visible.

\textbf{Bootstrapping a complete avatar from partial observations.} With large body regions never observed, there is no direct supervision for a significant portion of the avatar. We address this in three stages. First, we build a coarse 3DGS avatar from sparse multi-view images generated from the partially observed subject (Sec.~\ref{sec:coarse_avatar}). Second, we render this coarse avatar in structured motion sequences and refine the imperfect renderings via an identity-finetuned video diffusion model and diffusion inversion producing hallucinated supervision videos that cover previously unobserved body surfaces from multiple viewpoints and poses (Sec.~\ref{sec:hallucinated_supervision}). Third, we train a full avatar with pose-dependent deformation using these hallucinated videos as supervision (Sec.~\ref{sec:full_avatar}).

\textbf{Learning pose-dependent appearance without multi-view capture.} High-fidelity avatars require pose-dependent deformation to capture how clothing changes with body configuration. In the first stage, no multi-view supervision for pose-dependent deformation exists. We therefore learn an avatar that does not model pose-dependent deformations. We adapt the 2D Gaussian map design from Animatable Gaussians~\cite{li2024animatablegaussians} to learn a canonical pose-independent Gaussian map that captures the texture and structure of the human. We also adopt the use of per-identity networks that use 2D maps to represent both input pose and output 3DGS properties. To obtain the pose-dependent multi-view supervision missing from the original video, we design the structured motion sequences of the second stage (full-body turns paired with target poses such as sitting and reaching) to provide both pose-dependent deformations and in-effect 360$^\circ$ multi-view coverage per pose. In the third stage, these hallucinated videos supply the dense multi-view supervision needed to upgrade from canonical to pose-dependent Gaussian maps that capture high-fidelity appearance changes across body configurations. This canonical-to-pose-dependent progression is the architectural insight that makes pose-dependent learning possible (Secs.~\ref{sec:supervision_generation} and~\ref{sec:pose_dependent_maps}).

\textbf{Absorbing multi-view inconsistencies from diffusion-generated data.} Each hallucinated video is independently sampled from the diffusion model and is therefore not geometrically consistent across viewpoints and frames. Naively training on these videos produces artifacts. To address this, we decouple per-frame variations by introducing two distinct poses: a `map pose' and an `LBS pose.' Specifically, the map pose, which is taken as input for our per-identity networks, is shared across frames with similar poses to ensure consistent prediction of Gaussian attributes for a given pose. In contrast, the LBS pose, which deforms the output Gaussians from our per-identity networks, is optimized per-frame to absorb the geometric inconsistencies inherent in the video diffusion model (Sec.~\ref{sec:joint_optimization}).

\textbf{Preserving facial identity under unreliable diffusion outputs.} Video diffusion models generate identity-inconsistent faces. We partition the avatar into head and body regions via SMPL-to-FLAME correspondence, mask head pixels from the diffusion-based body supervision, and instead supervise the head through a dedicated multi-view face diffusion model that preserves identity (Sec.~\ref{sec:head_identity}).

We evaluate AHOY on YouTube videos featuring humans under significant occlusion and on multi-view capture data with comparable occlusion, and demonstrate state-of-the-art reconstruction quality. The resulting avatars are robust enough to be animated with novel poses and composited into 3DGS scenes reconstructed from casual cell-phone video, yielding an end-to-end pipeline from commodity RGB input to photorealistic human--scene rendering.

In summary, our contributions are:

\noindent 1) A method for reconstructing complete, animatable 3DGS avatars from in-the-wild monocular video under heavy occlusion, unlocking YouTube-scale footage as a source of 3D human assets.

\noindent 2) A hallucination-as-supervision approach using identity-finetuned video diffusion and diffusion inversion.

\noindent 3) A two-stage canonical-to-pose-dependent architecture that allows for learning high-fidelity pose-dependent deformations in 3DGS avatars.

\noindent 4) A map-pose/LBS-pose decoupling that enables pose-dependent Gaussian maps for high-fidelity avatar learning from multi-view-inconsistent generated data.

\noindent 5) A head/body split supervision for facial identity preservation.

\section{Related Work}

\noindent\textbf{Neural Rendering.}
NeRF~\cite{mildenhall2020nerf} and its extensions~\cite{mueller2022instant, barron2022mip, barron2023zipnerf, nerfstudio, nerf_review} enable high-quality novel-view synthesis but remain computationally expensive. 3DGS~\cite{kerbl2023gaussians} addresses this with explicit Gaussian primitives~\cite{lassner2021pulsar, westover1991phdsplatting} and has been extended to dynamic scenes~\cite{shaw2023swags, luiten2023dynamic, Wu2024CVPR, lee2024ex4dgs, li2023spacetime, mir2025gaspacho}, SLAM~\cite{keetha2024splatam}, mesh extraction~\cite{Huang2DGS2024, guedon2023sugar}, sparse-view reconstruction~\cite{mihajlovic2024SplatFields}, and avatar-scene composition~\cite{mir2026raga}. Beyond efficiency, the explicit nature of 3DGS makes it easy to edit, deform, and composite into new scenes, properties we rely on when animating reconstructed avatars and placing them in captured environments.

\noindent\textbf{Human Reconstruction and Neural Rendering.}
Dense body representations such as DensePose~\cite{guler2018densepose, neverova2018denseposetransfer, neverova2020cse} provide pixel-level surface correspondences for human understanding.
Mesh-based approaches~\cite{SMPL-X:2019, smpl2015loper, Bogo2016keepitsmpl, kanazawaHMR18, alldieck2018video, alldieck19cvpr, mir2024origin, mir20hps} and implicit functions~\cite{mescheder2019occupancy, park2019deepsdf, chen2021snarf, alldieck2021imghum, saito2020pifuhd, he2021arch++, huang2020arch, deng2020nasa, mir2020pix2surf} recover 3D humans but lack photorealism or reposability. NeRF-based methods~\cite{peng2021animatable, guo2023vid2avatar, weng_humannerf_2022_cvpr, jian2022neuman, haberman2023hdhumans, heminggaberman2024trihuman, li2022tava, liu2021neural, xu2021hnerf} produce photorealistic renderings from video, while 3DGS-based avatars~\cite{guo2025vid2avatarpro, kocabas2023hugs, moreau2024human, abdal2023gaussian, zielonka2023drivable, moon2024exavatar, li2024animatablegaussians, pang2024ash, lei2023gart, hu2024gaussianavatar, li2024relightableandanimatable, zheng2024gpsgaussian, jiang2024robust, dhamo2023headgas, qian20233dgsavatar, xu2024gaussian, junkawitsch2025eva} offer real-time rendering. None of these methods work with YouTube videos with significant occlusion.

\noindent\textbf{Humans from Monocular Input.}
Early implicit methods~\cite{saito2019pifu, zheng2020pamir, xiu2023econ, zhang2024sifu, huang2024tech} and optimization-based approaches~\cite{jiang2022selfrecon, hu2024gauhuman} reconstruct humans from single images or monocular video. Feed-forward models~\cite{qiu2025lhm, qiu2025pflhm, hong2023lrm, zhuang2025idol, sim2025persona} predict animatable avatars in seconds. Many methods leverage diffusion to synthesize multi-view images as an intermediate step~\cite{pan2025humansplat, peng2024charactergen, he2024magicman, qiu2025anigs, lu2025gas, chen2025synchuman, li2025pshuman, huang2025adahuman, buehler2026dreamliftanimate, dong2025moga, taubner2025cap4d}, while others extend to 4D or unconstrained settings~\cite{tan2025dressrecon, pang2024disco4d, shao2024human4dit, yang2024havefun, lin2025omnihuman}. All these methods assume a fully visible subject as input. Text-to-3D methods~\cite{poole2023dreamfusion, kolotouros2024dreamhuman, cao2024dreamavatar} generate humans from language prompts, but cannot reconstruct from visual observations. In contrast, our method recovers an animatable avatar of a particular person from occluded in-the-wild YouTube videos.

\noindent\textbf{Video Diffusion Models.}
Image and video diffusion models~\cite{Ho2020DDPM, Rombach2022LDM, saharia2022imagen, peebles2023dit, esser2024sd3, google2024imagen3, blattmann2023svd, yang2024cogvideox, wan2025wan, brooks2024sora, openai2025sora2, kong2025hunyuanvideo, seedance2025, wu2025hunyuanvideo15, seedance2025pro, seedance2026v2} have driven rapid progress in visual generation. A large body of work specializes these for human-centric synthesis, spanning pose-driven character animation~\cite{xu2024magicanimate, hu2024animateanyone, hu2025animateanyone2, guo2023animatediff, wang2024unianimate, wang2025unianimate_dit, gan2025humandit, zhou2025realisdance, luo2025dreamactor, men2024mimo, tu2025stableanimator, chang2023magicpose, xu2024anchorcrafter, zhu2024champ, zhang2024mimicmotion}, portrait animation~\cite{guo2024liveportrait, xu2024vasa, xie2024xportrait, wei2024aniportrait, xu2025hunyuanportrait, ma2024followyouremoji}, and video editing~\cite{jiang2025vace}, building on earlier warping-based methods~\cite{siarohin2019fomm, zhao2022tps}. LoRA-based personalization enables identity-specific finetuning of video diffusion models~\cite{abdal2025dynamicconcepts, abdal2025gridlora, dravid2024w2w}, while inversion techniques~\cite{abdal2019image2stylegan, abdal2020image2styleganpp, mokady2023nulltext, wang2025rfsolver, rout2025rfinversion} allow embedding real videos into the model's latent space for faithful reconstruction and editing. Video diffusion has also been used as supervision for 3DGS under sparse-view setups~\cite{zhong2025taming, nazarczuk2025vidar}. Unlike these works, we leverage video diffusion as a prior for 3DGS-based avatar reconstruction. Rather than using the diffusion model only to render novel views, we synthesize plausible appearance for body regions the input video never reveals and turn these generations into dense supervision for fitting the avatar.

\vspace{-4mm}

\section{Method}
\label{sec:method}

\begin{figure}[t]
  \centering
  \includegraphics[width=\linewidth]{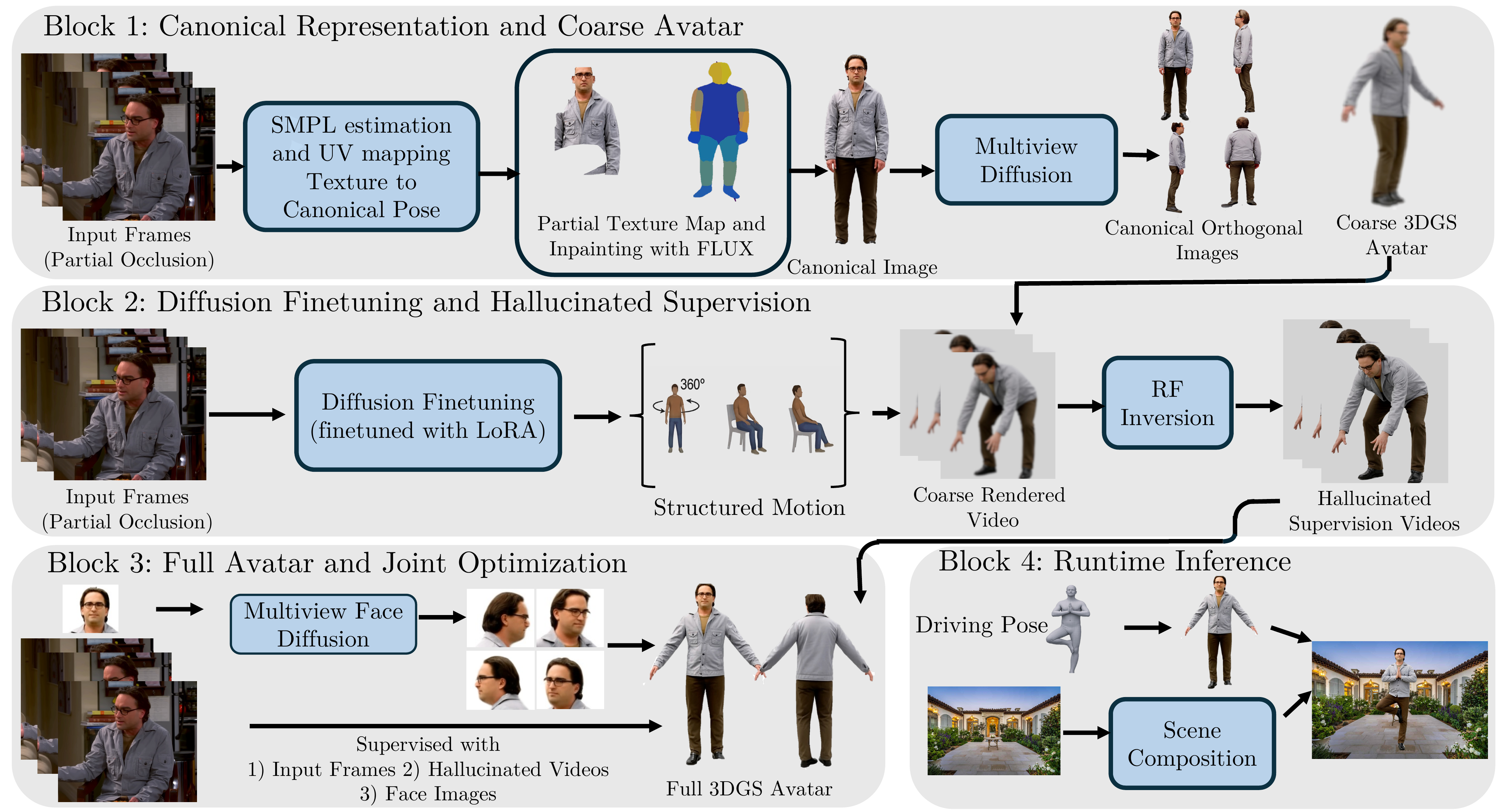}
  \caption{\textbf{Method overview.} \textbf{Block~1} (Sec.~\ref{sec:coarse_avatar}): We map observed textures from partially occluded video onto a canonical pose via DensePose UV correspondences, inpaint missing regions with FLUX, and generate multi-view canonical-pose images with multi-view diffusion to supervise a coarse 3DGS avatar using canonical Gaussian maps. \textbf{Block~2} (Sec.~\ref{sec:hallucinated_supervision}): We finetune a video diffusion model (Wan~2.2) with LoRA to capture the subject's identity, render the coarse avatar under structured motion sequences, and refine these renderings via RF-Inversion through the identity-finetuned latent space to produce hallucinated supervision videos. \textbf{Block~3} (Sec.~\ref{sec:full_avatar}): The hallucinated videos supervise a full avatar with pose-dependent Gaussian maps, where per-frame poses and cameras are jointly optimized to absorb multi-view inconsistencies; a separate FLAME-based head path preserves facial identity. \textbf{Block~4} (Sec.~\ref{sec:animation}): At inference, the avatar is driven by novel poses and composited into 3DGS scenes.}
  \label{fig:method}
  \vspace{-1mm}
\end{figure}
Given a monocular video in which a human subject is partially occluded, our goal is to reconstruct a complete, animatable 3D Gaussian avatar that can be driven by arbitrary poses. Our pipeline (Fig.~\ref{fig:method}) proceeds in four stages: (1)~coarse avatar construction from the partially observed subject (Sec.~\ref{sec:coarse_avatar}); (2)~hallucinated supervision generation via identity-finetuned video diffusion (Sec.~\ref{sec:hallucinated_supervision}); (3)~full avatar learning with pose-dependent Gaussian maps trained on this supervision (Sec.~\ref{sec:full_avatar}); and (4)~animation and scene composition (Sec.~\ref{sec:animation}).

\subsection{Coarse Avatar Construction}
\label{sec:coarse_avatar}

We first construct a coarse 3DGS avatar from the partially observed subject (Block~1 in Fig.~\ref{fig:method}).

\subsubsection{Texture mapping and inpainting.}
\label{sec:texture_mapping}
Given a monocular YouTube video with frames $\{\mat{I}_t\}_{t=1}^T$, we first estimate per-frame SMPL parameters $\{(\pose_t, \shape)\}$ using NLF~\cite{sarandi2024nlf}, and obtain visibility masks $\{\mat{M}_t\}$ via SAM~3~\cite{carion2025sam3}.

Using DensePose~\cite{guler2018densepose}, we assign each visible pixel a UV coordinate and accumulate them frame by frame into a canonical texture atlas with a first-write-wins rule to avoid temporal blending artifacts. We then define a fixed front-facing canonical pose $\pose_*$ and warp the atlas back to this pose via inverse DensePose correspondence~\cite{neverova2018denseposetransfer}, producing a partial canonical-pose image $\mat{I}_*^{\mathrm{part}}$. Regions that were never observed remain empty. A pretrained FLUX~\cite{esser2024sd3} inpainting model fills in these missing regions, yielding a complete canonical-pose image $\mat{I}_*$ that serves as the basis for multi-view generation. (Full details in the supplementary.)

\subsubsection{Multi-view generation and coarse 3DGS avatar.}
\label{sec:multiview_coarse}
The inpainted canonical image $\mat{I}_*$ provides only a single viewpoint. To obtain broader coverage, we generate $V\!-\!1\!=\!3$ additional canonical-pose views (back, left, right) using a pretrained multi-view diffusion model~\cite{chen2025synchuman}, discarding its 3D reconstruction branch. Together with the frontal view $\mat{I}_*^1\!=\!\mat{I}_*$, this yields $V\!=\!4$ canonical-pose images $\{\mat{I}_*^v\}_{v=1}^V$ at known orthogonal cameras $\{\camerapose_v\}_{v=1}^V$.

\textbf{Canonical Gaussian maps.}
Following~\cite{mir2025gaspacho, li2024animatablegaussians}, we densely sample $N_\Hum$ points on the SMPL mesh in canonical pose $\pose_*$ to define a canonical Gaussian template $\set{G}^{\Hum}_*$. These are projected from front and back views to create canonical position maps with one-to-one pixel-to-Gaussian correspondence. Unlike the pose-dependent human maps in~\cite{mir2025gaspacho}, we use time-invariant canonical maps, as our monocular occluded setting lacks the multi-view coverage needed for pose-conditioned offsets.

\textbf{StyleUNet prediction.}
A StyleUNet~\cite{wang2023styleavatar} $S$ processes the canonical maps and predicts per-pixel offsets for each Gaussian property from the template, yielding the canonical Gaussian set $\set{G}^{\Hum}\!=\!S(\pose_*)$. Additionally, $S$ predicts per-Gaussian skinning weights $\gskin^{\Hum}_{k,j}$ for $j\!=\!1,\ldots,J$ joints.

\textbf{LBS deformation.}
For each frame $t$ with estimated pose $\pose_t$, we deform the canonical Gaussians into posed space via Linear Blend Skinning (LBS) using the predicted skinning weights $\gskin^{\Hum}_{k,j}$. We denote the complete deformation as $\deform(\set{G}^{\Hum};\pose_t)$; full details are provided in the supplementary material. Colors and opacities remain as predicted by the StyleUNet.

\textbf{Training objective.}
The coarse avatar is supervised by two sources: (i)~the $V$ canonical-pose images at known cameras, and (ii)~the $T$ observed video frames at estimated poses and observed camera $\camerapose^{\mathrm{obs}}$ (the camera associated with the NLF reference frame). Let $\hat{\mat{I}}_t\!=\!\rasterize(\deform(\set{G}^{\Hum};\pose_t);\camerapose^{\mathrm{obs}})$ denote the rendering at frame $t$. The training loss is:
\begin{equation}
\Loss_{\mathrm{coarse}} = \sum_{v=1}^{V} \| \mat{I}_*^v - \rasterize(\set{G}^{\Hum}; \camerapose_v) \|_1
  \;+\; \sum_{t=1}^{T} \| (\mat{I}_t - \hat{\mat{I}}_t) \had \mat{M}_t \|_1,
\label{eq:loss_coarse}
\end{equation}
where $\had$ denotes element-wise multiplication and $\mat{M}_t$ masks out occluded regions. We additionally employ a perceptual loss~\cite{zhang2018lpips} with the same masking (see supplementary). This yields a coarse avatar faithful to the visible portions of the subject but with artifacts in unobserved regions.

\subsection{Hallucinated Supervision Generation}
\label{sec:hallucinated_supervision}

The coarse avatar (Sec.~\ref{sec:coarse_avatar}) is supervised by only $V\!=\!4$ canonical-pose images and the occluded input frames. This limited coverage means that unobserved regions (the back of the body, inner arms, or any area persistently hidden by the occluder) remain poorly reconstructed, and the avatar cannot generalize to novel poses and camera angles beyond those seen during training. We now proceed to generate hallucinated videos for further supervision (Block~2 in Fig.~\ref{fig:method}).

\subsubsection{Identity-specific video diffusion finetuning.}
\label{sec:identity_finetuning}
A generic video diffusion model lacks knowledge of the subject's specific appearance, so we finetune one on the input video to build an identity-specific prior. We adapt the identity encoding stage of Dynamic Concepts~\cite{abdal2025dynamicconcepts} to Wan~2.2~\cite{wan2025wan}. Let $\vec{v}_{\phi}$ denote the velocity field of the pretrained rectified flow model. We learn a LoRA update $\mat{A}\mat{B}$ together with a learnable text token $[\mathrm{v}]$ that binds to the target identity in the text encoder's embedding space. Following~\cite{abdal2025dynamicconcepts}, we treat the input video as an \emph{unordered} set of frames (removing temporal information) so that the LoRA captures appearance and identity without entangling motion dynamics. The training objective is the flow matching loss:
\begin{equation}
\Loss_{\mathrm{id}} = \mathbb{E}_{\tau,\,\vec{z}_0}\!\left[\,\bigl\| (\vec{z}_1 - \vec{z}_0) - \vec{v}_{\phi}(\vec{z}_\tau,\,\tau;\,\text{``a}\;[\mathrm{v}]\;\text{person''}) \bigr\|^2 \,\right],
\label{eq:lora_loss}
\end{equation}
where $\vec{z}_0\!=\!\func{Enc}(\mat{I}_t)$ is the VAE-encoded latent, $\vec{z}_1\!\sim\!\Normal(\vec{0},\mathbb{I})$ is noise ($\mathbb{I}$ denotes the identity matrix), $\vec{z}_\tau\!=\!(1\!-\!\tau)\vec{z}_0 + \tau\vec{z}_1$ is the interpolated latent at flow time $\tau\!\in\![0,1]$, and the velocity field is parameterized by the LoRA update $\mat{W}'\!=\!\mat{W} + \mat{A}\mat{B}$ applied to the pretrained weights $\mat{W}$. After training, the finetuned model $\vec{v}_\phi$ generates videos that faithfully preserve the subject's identity, appearance, and clothing, which we exploit in the next stages to hallucinate occluded regions.

\subsubsection{Supervision video generation.}
\label{sec:supervision_generation}
We now use the identity-finetuned model to generate supervision videos that cover the missing body regions. Rather than generating arbitrary motions, we curate a set of $N$ motion sequences designed to systematically expose all body surfaces to the camera while providing the multi-view coverage needed to learn pose-dependent Gaussian maps (Sec.~\ref{sec:full_avatar}).

\textbf{Structured Motion (turn).}
Each turn-based sequence pairs a target body action (\eg, arms lifting or both arms raised) with a full 360$^\circ$ body rotation. The key observation is that all frames within such a sequence can be treated as views of the \emph{same} pose from different angles: the body action remains approximately constant while the rotation sweeps the viewpoint around the person. A turn with $T'$ frames thus provides in-effect $T'$-view supervision of a single pose, enabling the model to learn how the Gaussian representation should look from every direction under that specific body configuration.

\textbf{Structured Motion (static).}
Certain poses, such as sitting, are incompatible with a full-body turn. For these, we generate multiple videos of the same sitting pose from different camera angles. We then synchronize frames across these videos where the pose matches and supervise each pose with the images from all corresponding videos, providing cross-video multi-view supervision analogous to the within-video coverage of the turn sequences.

\textbf{Coarse rendering.}
For each motion sequence $n$, we animate the coarse avatar with the corresponding poses $\{\pose'_t\}_{t=1}^{T'}$ and render from the observed camera $\camerapose^{\mathrm{obs}}$:
\begin{equation}
\hat{\mat{V}}_n = \bigl\{ \rasterize\!\bigl(\deform(\set{G}^{\Hum};\pose'_t);\,\camerapose^{\mathrm{obs}}\bigr) \bigr\}_{t=1}^{T'}.
\label{eq:coarse_render}
\end{equation}
These renderings capture the correct body pose and silhouette but contain artifacts and missing textures in regions that were occluded in the input video. We refine them through the diffusion latent space below.

\subsubsection{Hallucination via RF-Inversion.}
\label{sec:rf_inversion}
The coarse renderings $\hat{\mat{V}}_n$ from above preserve the target body layout but suffer from texture artifacts in unobserved regions. We refine them by passing each video through the latent space of the identity-finetuned model $\vec{v}_\phi$, leveraging RF-Inversion~\cite{rout2025rfinversion} to replace artifacts with identity-consistent appearance.

Let $\func{Enc}$ and $\func{Dec}$ denote the VAE encoder and decoder, and let $\vec{z}_0\!=\!\func{Enc}(\hat{\mat{V}}_n)$ be the encoded coarse video. The forward rectified flow ODE maps this to a noisy latent $\vec{z}_1$, and reversing the flow with the identity-finetuned model decodes a hallucinated video:
\begin{equation}
\vec{z}_1 = \vec{z}_0 + \!\int_0^1 \vec{v}_\phi(\vec{z}_\tau,\tau)\,\dd\tau, \qquad
\tilde{\vec{z}}_0 = \vec{z}_1 + \!\int_1^0 \vec{v}_\phi(\vec{z}_\tau,\tau)\,\dd\tau, \qquad
\tilde{\mat{V}}_n = \func{Dec}(\tilde{\vec{z}}_0).
\label{eq:rf_inversion}
\end{equation}
The \emph{forward pass encodes the spatial structure} (body pose and silhouette) of the coarse rendering into $\vec{z}_1$, while the \emph{reverse pass preserves this layout but steers the appearance} toward the subject's identity via the LoRA, filling in previously missing regions with plausible textures. Applying this process to all $N$ motion sequences yields the hallucinated video set $\{\tilde{\mat{V}}_n\}_{n=1}^N$, which provides multi-view body coverage that serves as supervision for the full avatar (Sec.~\ref{sec:full_avatar}).

\subsection{Learning the Full Avatar under Multi-View Inconsistencies}
\label{sec:full_avatar}

The hallucinated videos $\{\tilde{\mat{V}}_n\}_{n=1}^N$ from Sec.~\ref{sec:hallucinated_supervision} provide dense body coverage but are not perfectly multi-view consistent, as each video is independently sampled from the diffusion model. We address this with (i)~a switch from canonical to pose-dependent Gaussian maps, enabled by the in-effect multi-view supervision from the hallucinated sequences, (ii)~joint optimization of Gaussian, pose, and camera parameters to absorb inconsistencies, and (iii)~a separate FLAME-based head supervision path that preserves facial identity.

\subsubsection{Pose-dependent Gaussian maps.}
\label{sec:pose_dependent_maps}
The coarse stage (Sec.~\ref{sec:coarse_avatar}) used time-invariant canonical maps because multi-view supervision for multiple poses was unavailable. The hallucinated videos now provide this coverage: in a turn-based sequence, all $T'$ frames observe approximately the same body pose from different viewpoints, yielding in-effect $T'$-view supervision for that pose. For static-pose sequences (\eg, sitting), we synchronize frames across videos rendered from different angles where the same pose occurs, providing cross-video multi-view supervision. This dense coverage enables us to upgrade to pose-dependent Gaussian maps, following~\cite{mir2025gaspacho, li2024animatablegaussians}: we project posed SMPL vertices onto front and back UV maps for each frame, and a re-initialized StyleUNet $S$ takes these pose-dependent maps as input and predicts per-pixel Gaussian property offsets from the canonical template, yielding $\set{G}^{\Hum}_n\!=\!S(\pose_n^{\mathrm{map}})$.

A crucial distinction arises between two roles of the pose: the \emph{map pose} $\pose_n^{\mathrm{map}}$, which is the input to the StyleUNet and remains \emph{fixed} across all frames of sequence $n$, and the \emph{LBS pose} $\pose_{n,t}^{\mathrm{lbs}}$, which deforms the predicted Gaussians into posed space and varies per frame. For a turn-based sequence, the map pose encodes the target body configuration (\eg, arms raised) so that $S$ produces a single set of Gaussians representing that pose; the per-frame LBS pose then handles the body rotation, mapping these Gaussians to the correct 3D orientation for each viewpoint.

\subsubsection{Joint optimization with pose and camera correction.}
\label{sec:joint_optimization}
Since the hallucinated videos are occlusion-free, we estimate SMPL poses with NLF~\cite{sarandi2024nlf} and assume cameras in the NLF reference frame. These estimates initialize both the map pose $\pose_n^{\mathrm{map}}$ (used to create pose-dependent input maps for $S$) and the per-frame LBS pose $\pose_{n,t}^{\mathrm{lbs}}$.
We optimize $S$ jointly with per-frame LBS pose corrections $\{\Delta\pose_{n,t}\}$ and camera corrections $\{\Delta\camerapose_{n,t}\}$ for each hallucinated frame. Let $\tilde{\mat{I}}_{n,t}$ denote frame $t$ of the $n$-th hallucinated video. We exclude head pixels using a segmentation mask $\mat{M}_{n,t}^{\mathrm{head}}$ obtained from Sapiens~\cite{khirodkar2024sapiens} and SAM~3~\cite{carion2025sam3}, since diffusion-generated faces are unreliable for identity preservation. The body supervision loss is:
\begin{equation}
\Loss_{\mathrm{body}} = \sum_{n,t} \bigl\| \bigl(\tilde{\mat{I}}_{n,t} - \hat{\mat{I}}_{n,t}\bigr) \had \bigl(1 - \mat{M}_{n,t}^{\mathrm{head}}\bigr) \bigr\|_1,
\label{eq:loss_body}
\end{equation}
where $\hat{\mat{I}}_{n,t}\!=\!\rasterize\!\bigl(\deform(\set{G}^{\Hum}_n;\,\pose_{n,t}^{\mathrm{lbs}}\!+\!\Delta\pose_{n,t});\,\camerapose^{\mathrm{obs}}\!+\!\Delta\camerapose_{n,t}\bigr)$ is the rendered image with corrected LBS pose and camera. The map pose $\pose_n^{\mathrm{map}}$ is shared across all frames of sequence $n$, so every hallucinated view supervises the same pose-dependent Gaussian map. The LBS pose corrections $\Delta\pose_{n,t}$ absorb per-frame multi-view inconsistencies from the diffusion-generated data, introducing a deliberate divergence between the map pose and the LBS pose that allows the canonical Gaussians to remain consistent while the deformation adapts to each frame.

We retain observed-frame supervision $\Loss_{\mathrm{obs}} = \sum_{t} \| (\mat{I}_t - \hat{\mat{I}}_t^{\mathrm{obs}}) \had \mat{M}_t \|_1$ from the input video, analogous to the second term of Eq.~\ref{eq:loss_coarse}, where $\hat{\mat{I}}_t^{\mathrm{obs}}$ is rendered with the original estimated poses and cameras.

\subsubsection{Head identity preservation via FLAME.}
\label{sec:head_identity}
Head regions in the Wan-generated videos are unreliable for facial identity. Instead, we leverage the multi-view face diffusion model from CAP4D~\cite{taubner2025cap4d} to generate identity-consistent head images $\{\mat{I}^{\mathrm{face}}_p\}_{p=1}^P$ from $P$ viewpoints with known cameras $\{\camerapose_p^{\mathrm{face}}\}$.

Using the SMPL-to-FLAME~\cite{li2017flame} vertex correspondence, we partition the canonical Gaussians into head and body regions:
\begin{equation}
\set{G}^{\mathrm{head}} = \bigl\{ (\gmean_k, \ldots) \in \set{G}^{\Hum} \;\big|\; k \in \set{H}_{\mathrm{head}} \bigr\},
\label{eq:head_partition}
\end{equation}
where $\set{H}_{\mathrm{head}}\!\subset\!\{1,\ldots,N_\Hum\}$ is the set of Gaussian indices corresponding to the SMPL head region. For head supervision, we deform only these Gaussians using FLAME expression parameters $\expr_p$:
\begin{equation}
\Loss_{\mathrm{head}} = \sum_{p=1}^{P} \bigl\| \mat{I}^{\mathrm{face}}_p - \rasterize\!\bigl(\deform_{\mathrm{FLAME}}(\set{G}^{\mathrm{head}};\,\expr_p);\,\camerapose_p^{\mathrm{face}}\bigr) \bigr\|_1,
\label{eq:loss_head}
\end{equation}
where $\deform_{\mathrm{FLAME}}$ applies FLAME-based deformation to the head Gaussians while leaving their canonical positions shared with the full body representation.

\subsubsection{Total objective.}
\label{sec:total_objective}
The full avatar is trained with the combined loss:
\begin{equation}
\Loss = \lambda_{\mathrm{body}} \Loss_{\mathrm{body}} + \lambda_{\mathrm{obs}} \Loss_{\mathrm{obs}} + \lambda_{\mathrm{head}} \Loss_{\mathrm{head}} + \lambda_{\mathrm{per}} \Loss_{\mathrm{per}} + \lambda_{\mathrm{reg}} \Loss_{\mathrm{reg}},
\label{eq:loss_full}
\end{equation}
where $\Loss_{\mathrm{per}}$ is a perceptual loss~\cite{zhang2018lpips} and $\Loss_{\mathrm{reg}}$ regularizes the skinning weights toward SMPL-derived values. The weighting coefficients $\lambda_{\mathrm{body}}$, $\lambda_{\mathrm{obs}}$, $\lambda_{\mathrm{head}}$, $\lambda_{\mathrm{per}}$, and $\lambda_{\mathrm{reg}}$ balance the respective terms. Because the hallucinated sequences span a wide range of body configurations (standing, sitting, arms raised, turning), the training pose distribution is broad enough that the learned pose-dependent Gaussian maps generalize well to novel driving poses at inference time.

\newcommand{\compinw}{0.07\linewidth}
\newcommand{\compmw}{0.305\linewidth}
\newcommand{\comph}{2.3cm}
\newcommand{\comphocc}{2.3cm}

\begin{figure*}[t]
    \centering
    \setlength{\tabcolsep}{1pt}
    \renewcommand{\arraystretch}{0.6}
    \begin{tabular}{cccc}
        {\tiny Input} &
        {\tiny \textbf{Ours}} &
        {\tiny LHM~\cite{qiu2025lhm}} &
        {\tiny IDOL~\cite{zhuang2025idol}} \\[2pt]
        \includegraphics[width=\compinw,height=\comph,keepaspectratio]{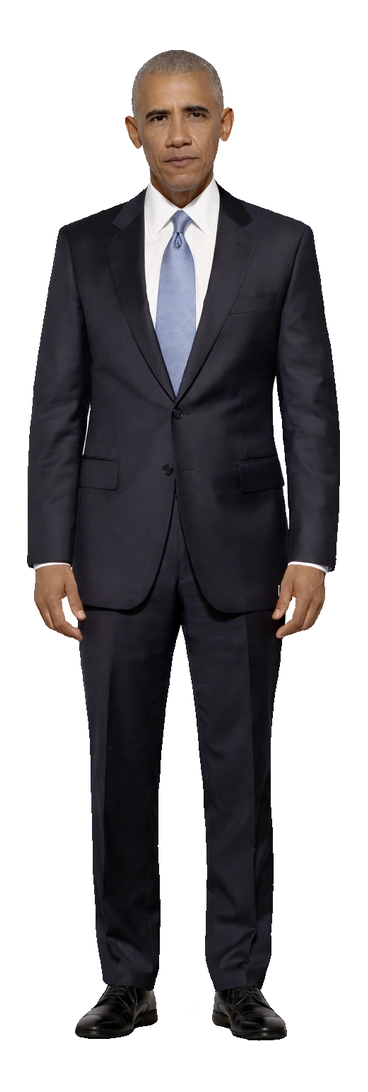} &
        \includegraphics[width=\compmw,height=\comph,keepaspectratio]{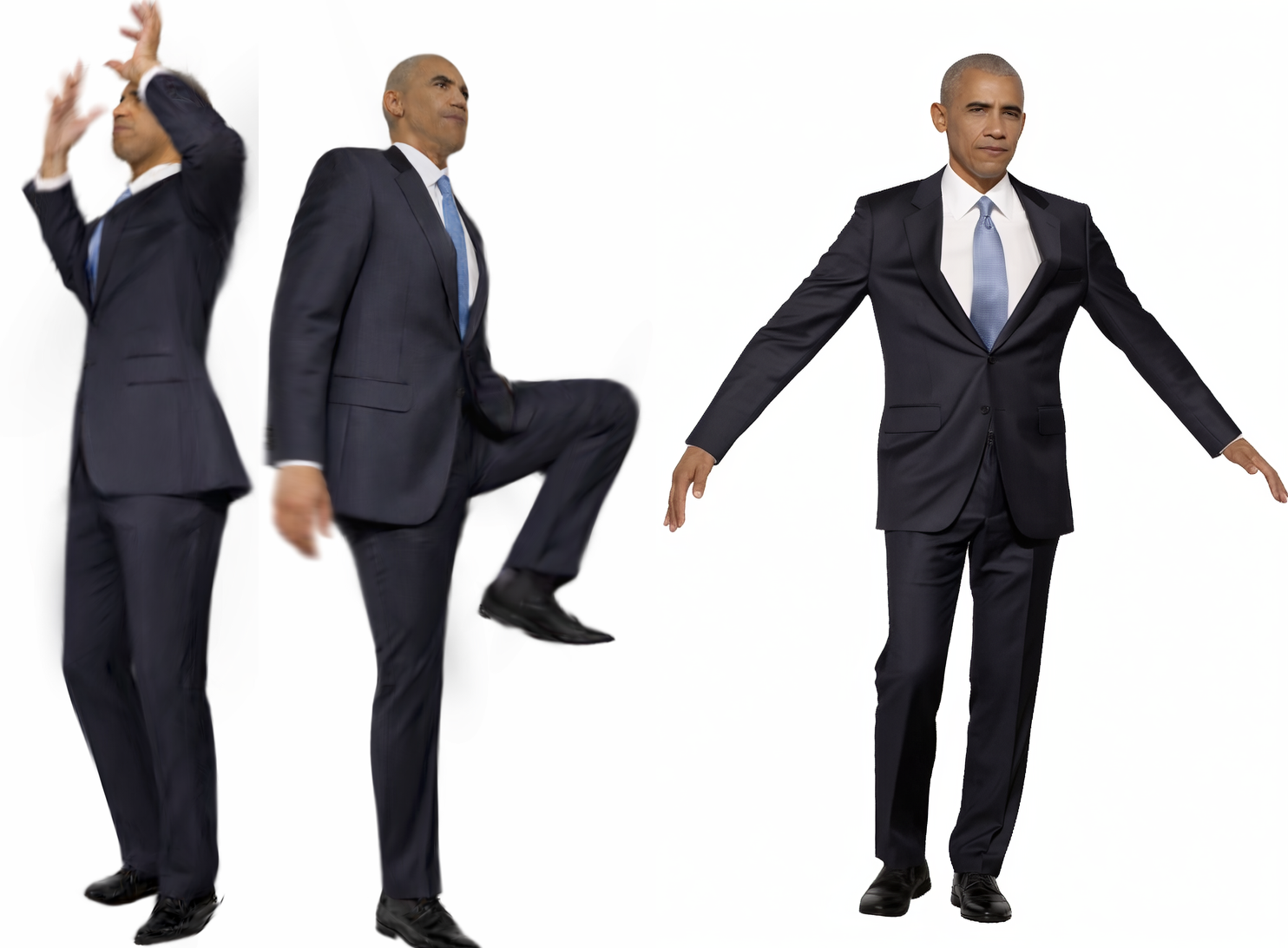} &
        \includegraphics[width=\compmw,height=\comph,keepaspectratio]{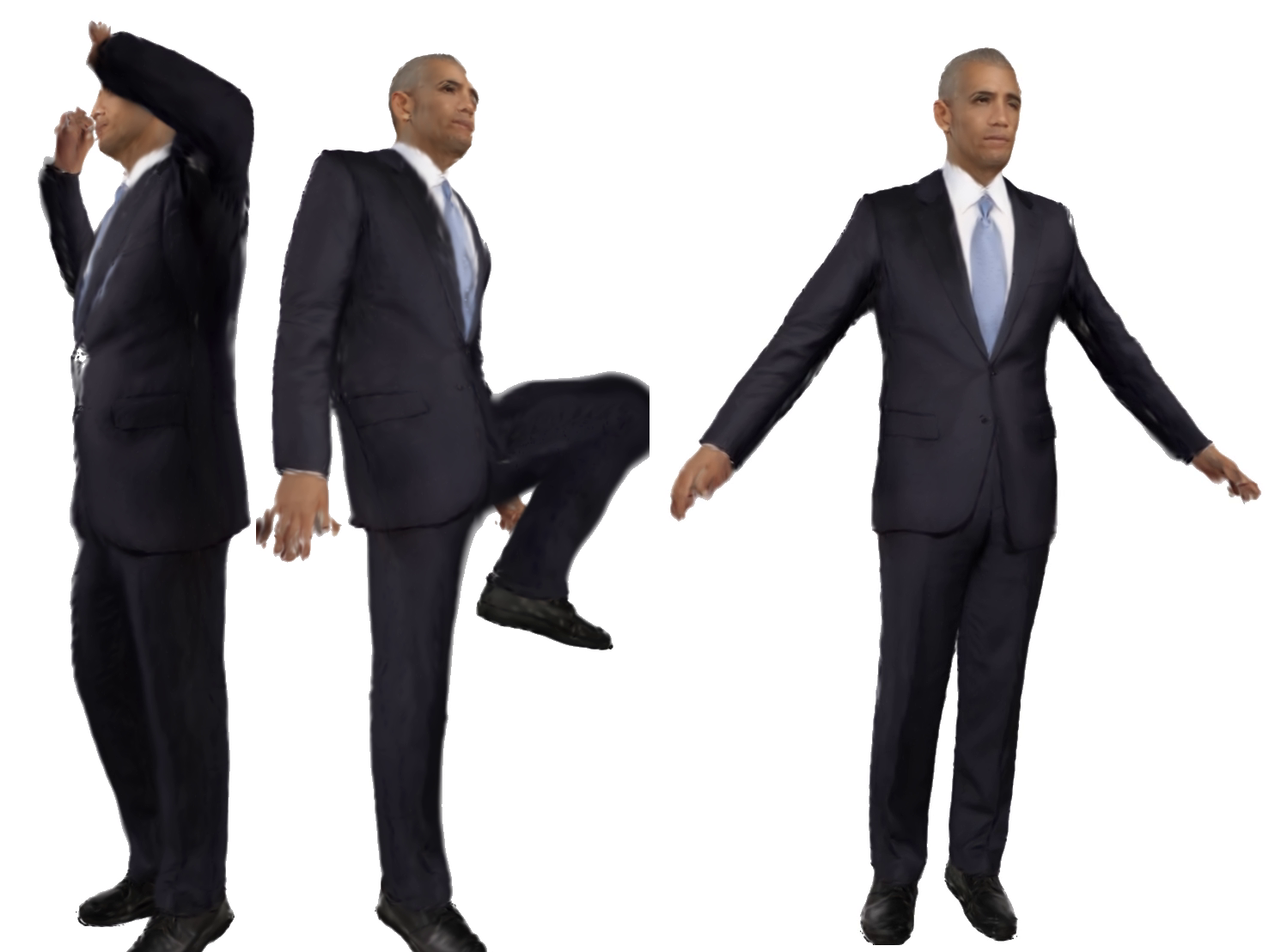} &
        \includegraphics[width=\compmw,height=\comph,keepaspectratio]{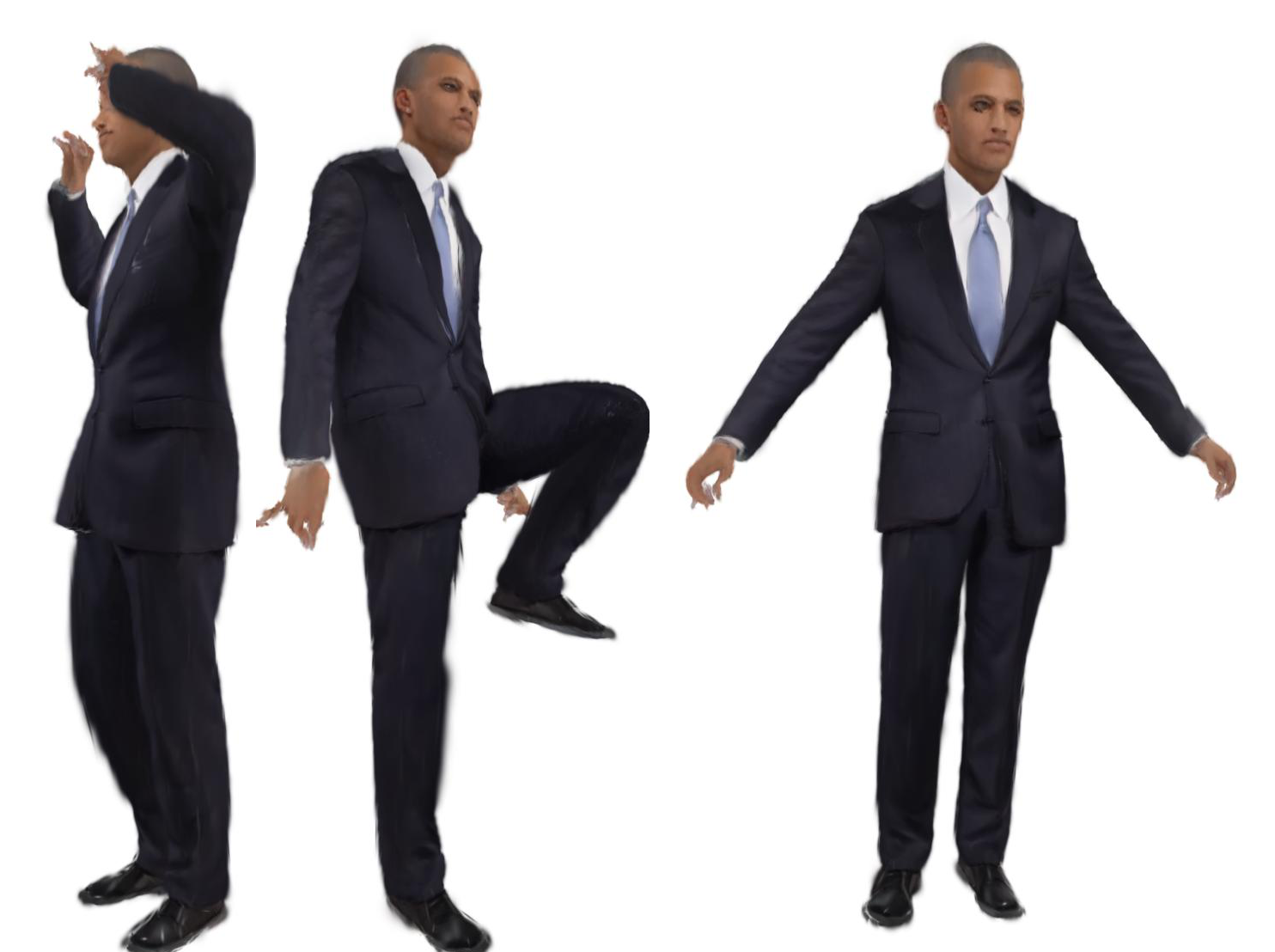} \\[1pt]
        \includegraphics[width=\compinw,height=\comph,keepaspectratio]{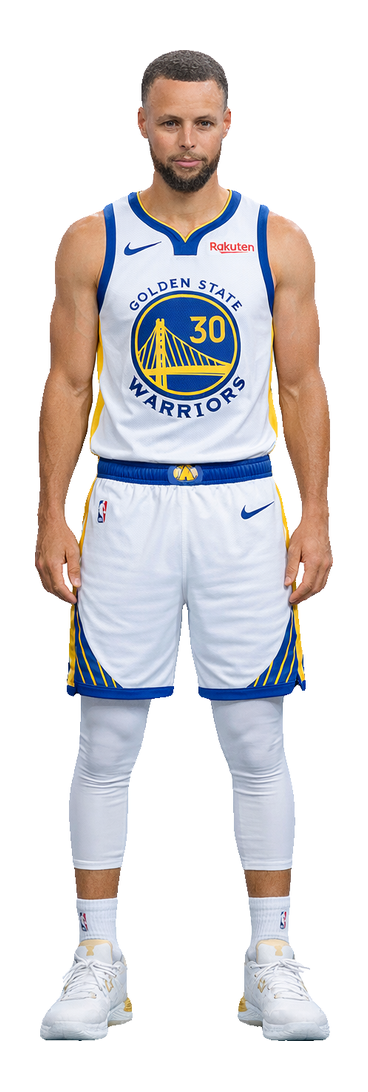} &
        \includegraphics[width=\compmw,height=\comph,keepaspectratio]{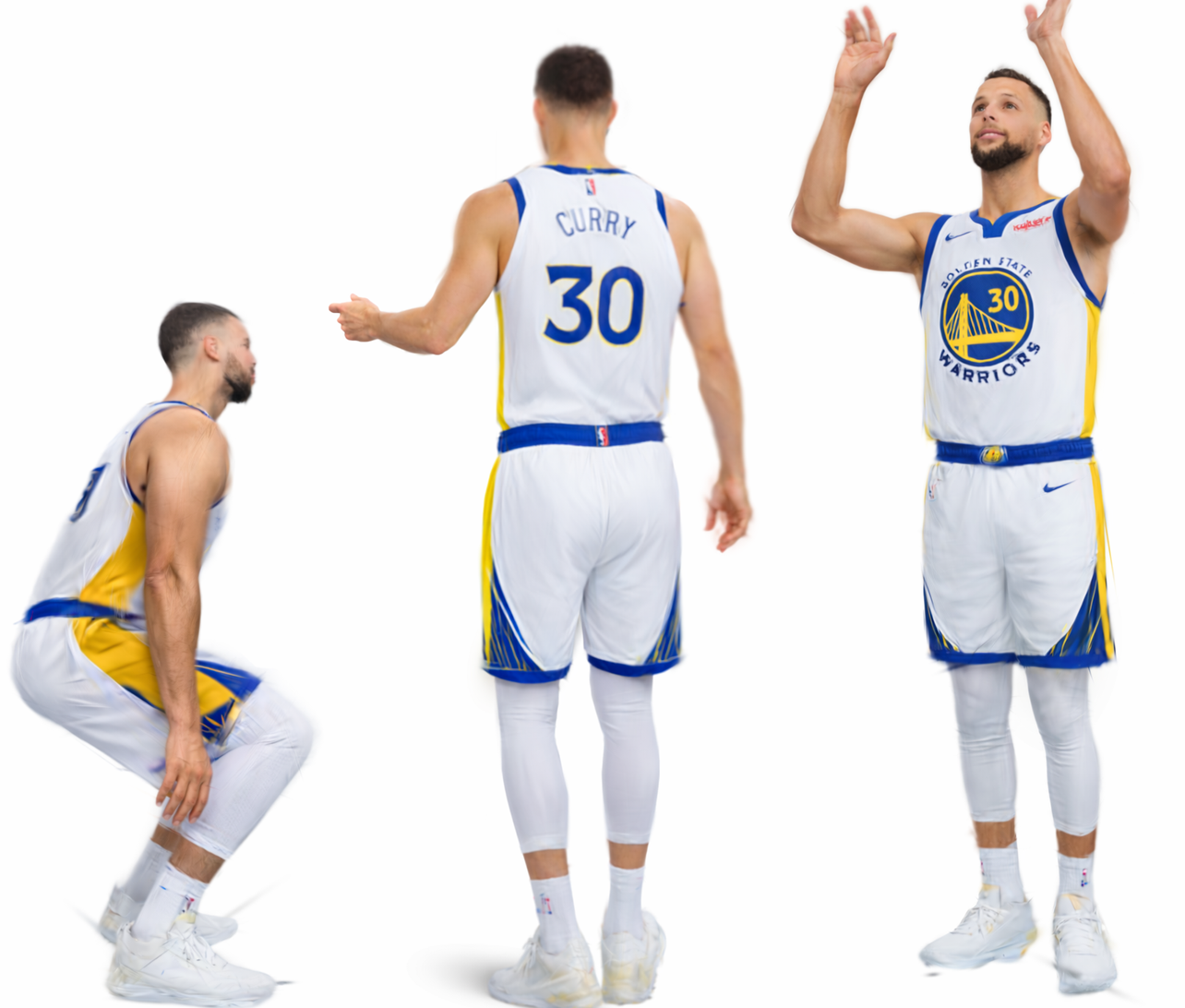} &
        \includegraphics[width=\compmw,height=\comph,keepaspectratio]{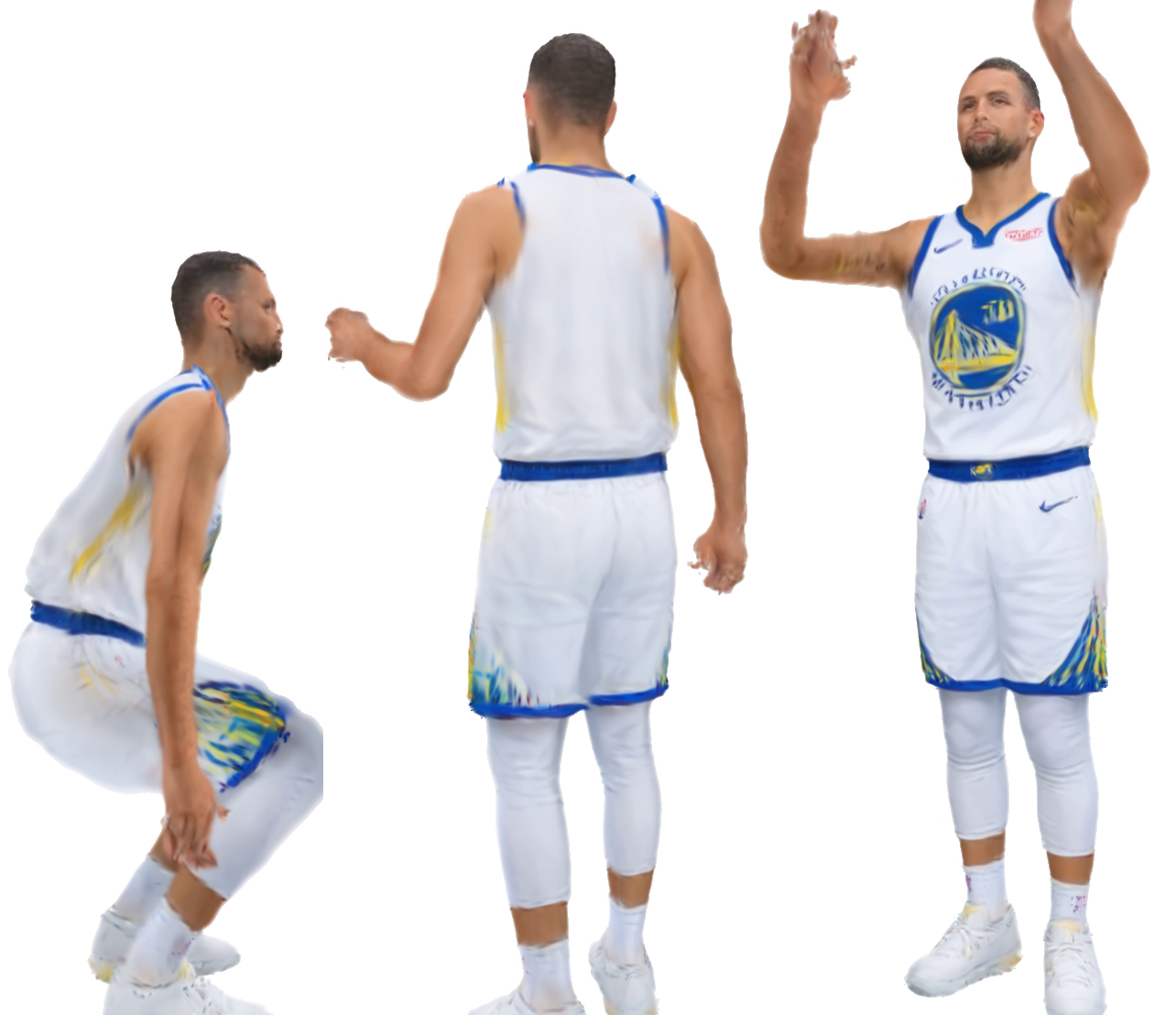} &
        \includegraphics[width=\compmw,height=\comph,keepaspectratio]{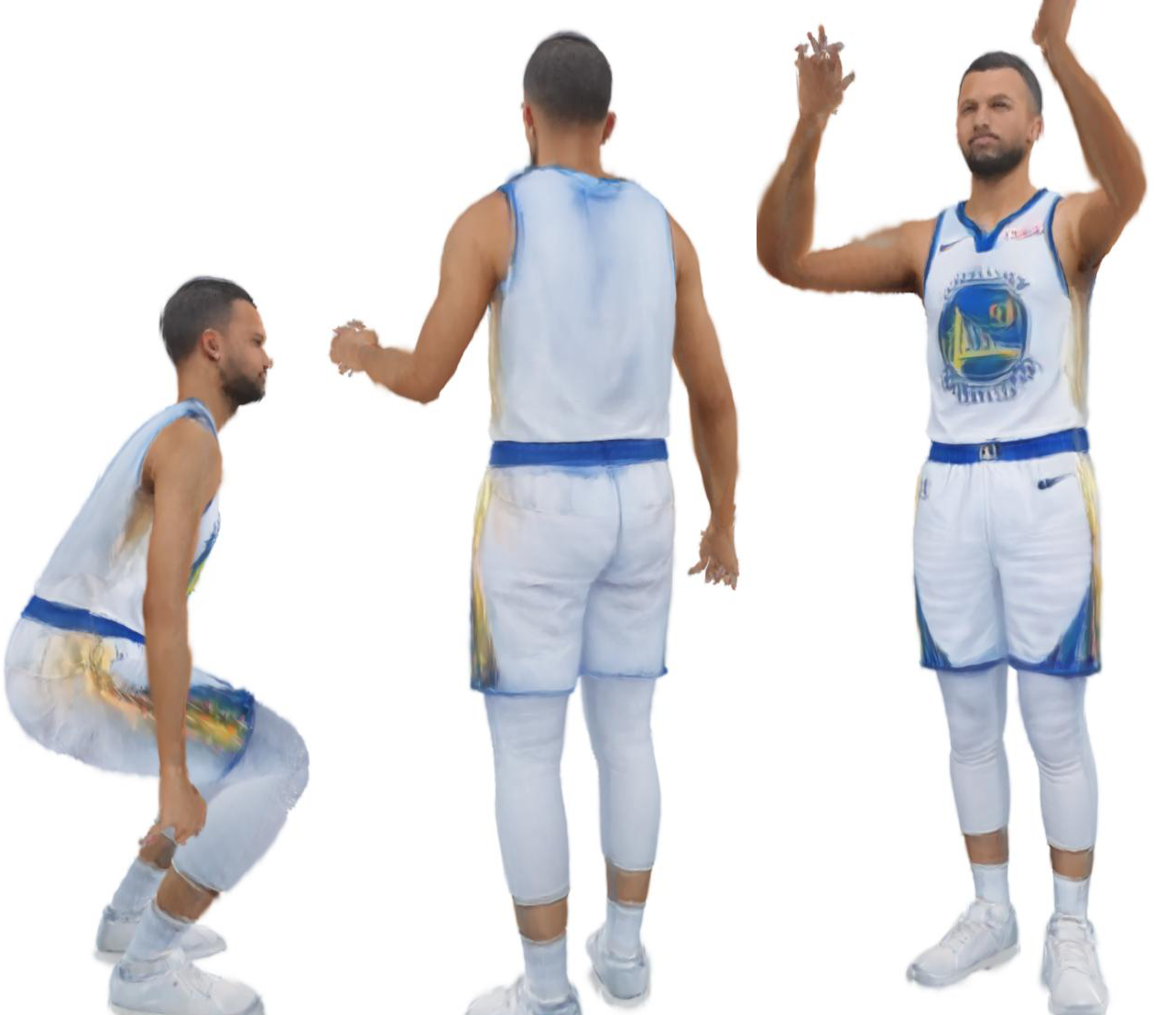} \\[-3pt]
        \multicolumn{4}{c}{\rule{0.95\linewidth}{0.4pt}} \\[-1pt]
        \includegraphics[height=\comphocc]{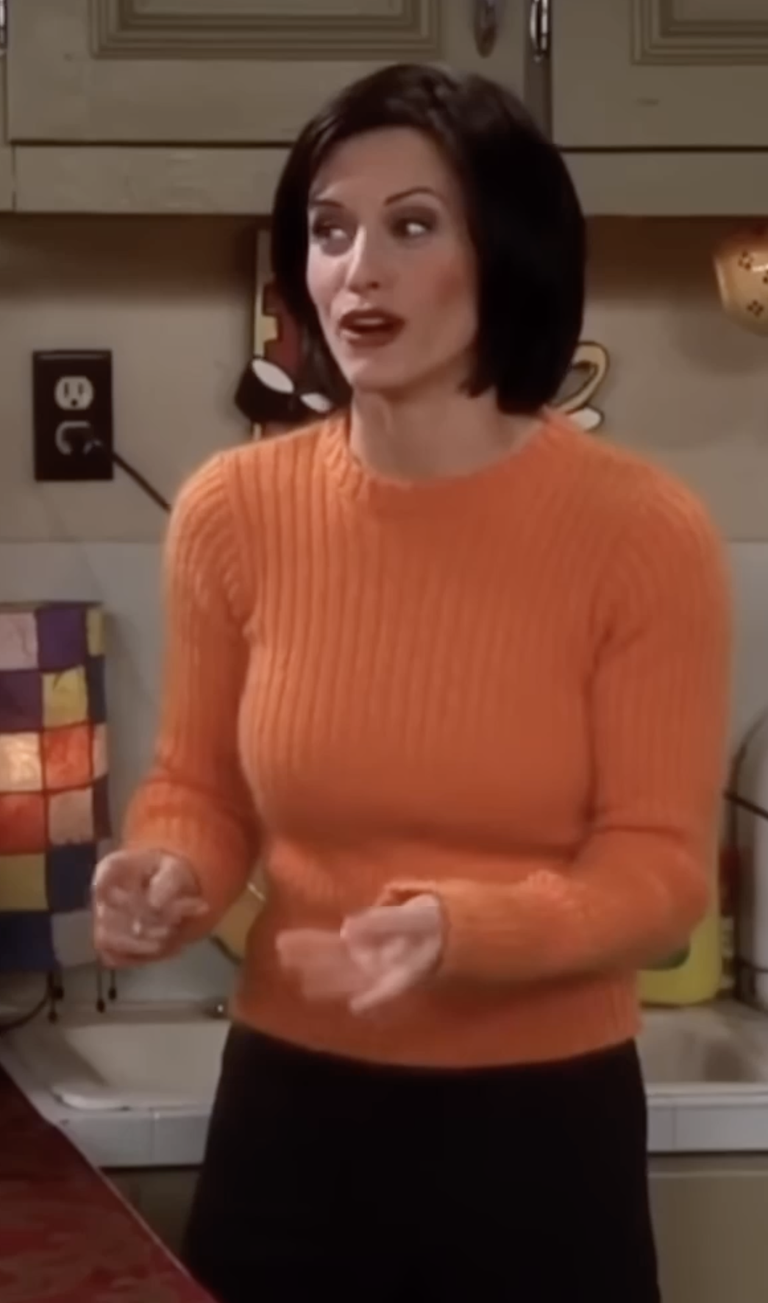} &
        \includegraphics[width=\compmw,height=\comphocc,keepaspectratio]{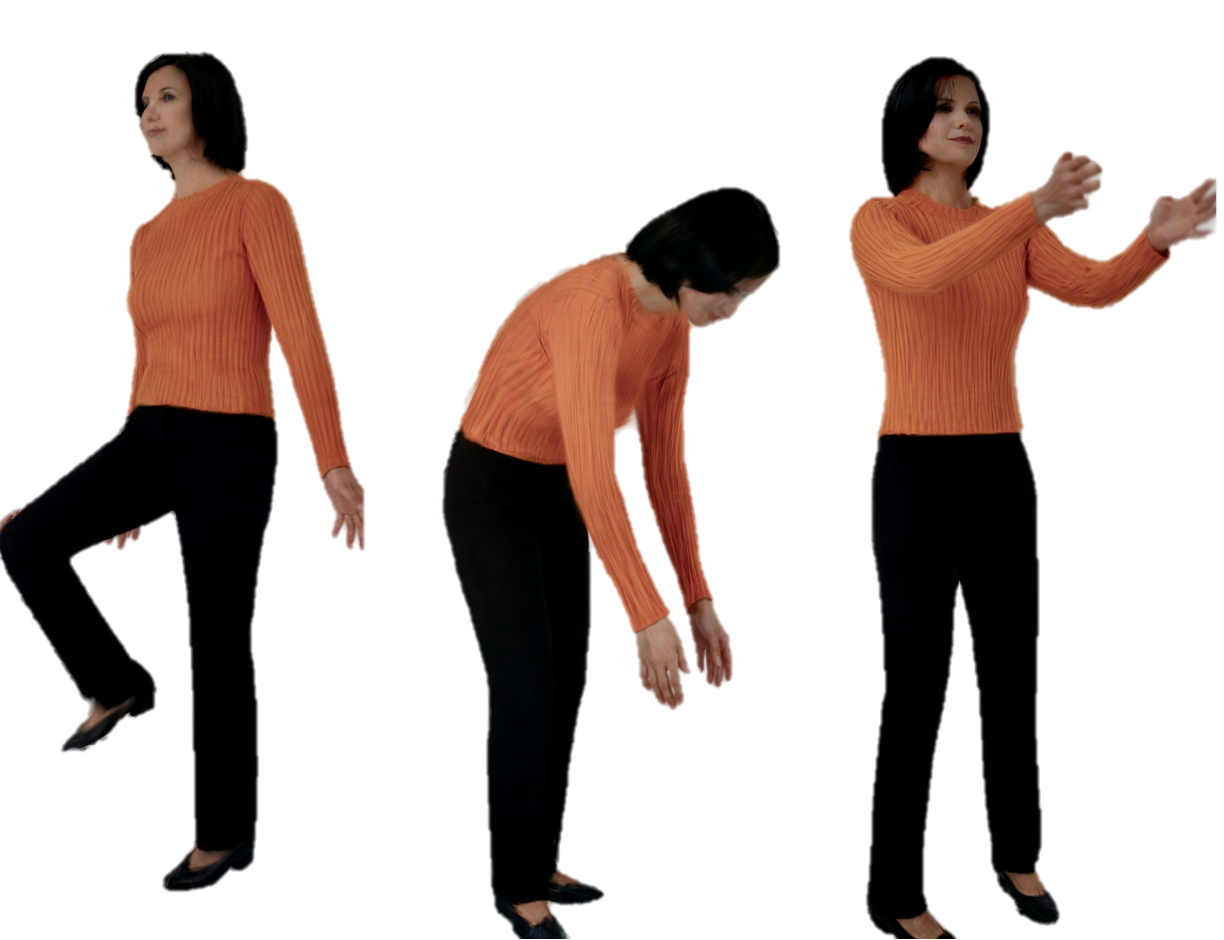} &
        \includegraphics[width=\compmw,height=\comphocc,keepaspectratio]{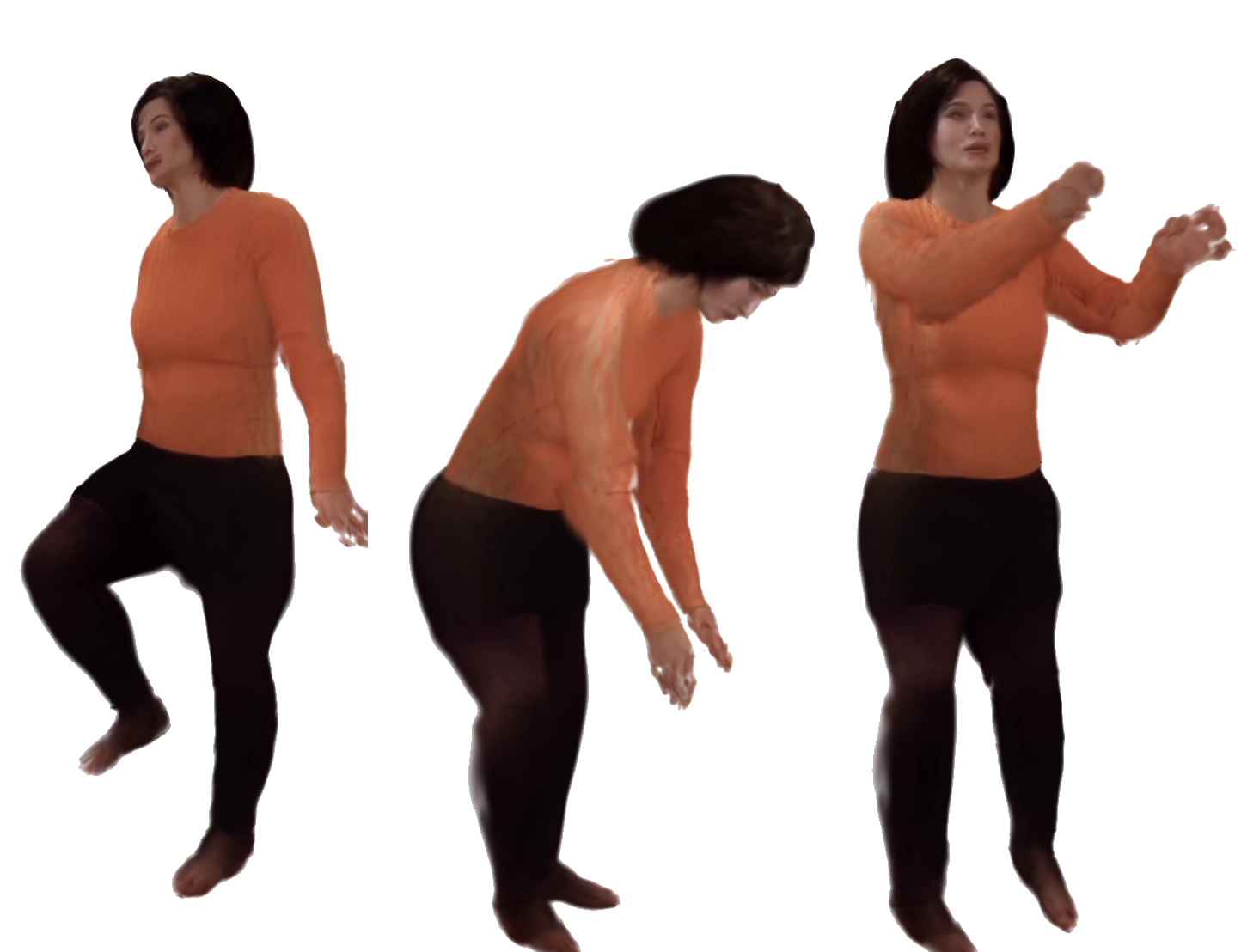} &
        \includegraphics[width=\compmw,height=\comphocc,keepaspectratio]{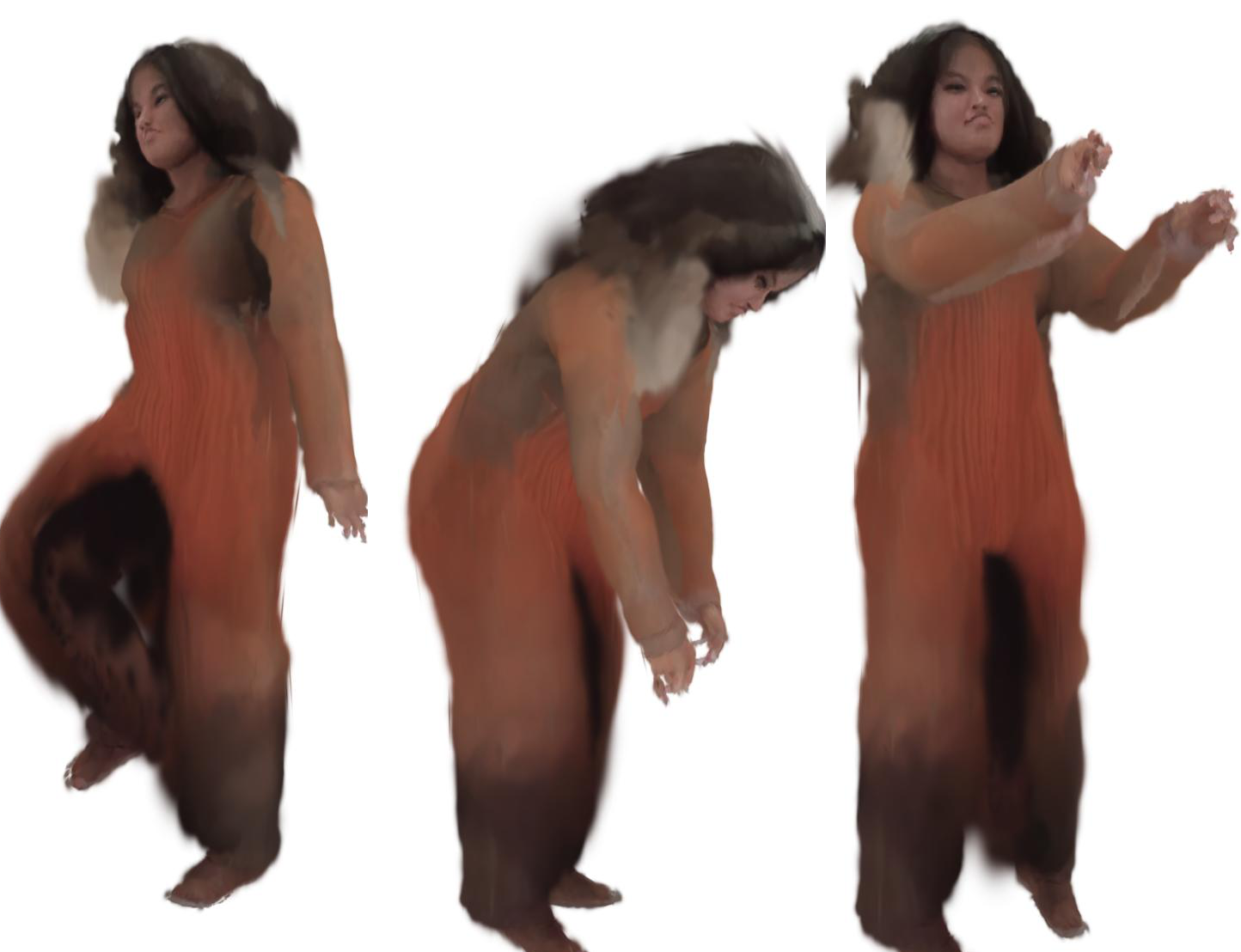} \\
    \end{tabular}
    \caption{\textbf{Animation comparison (Zoom In).} Top: canonical-pose input. Bottom (below line): occluded input. Ours produces higher-fidelity avatars in both settings.}
    \label{fig:comp_anim}
\vspace{-6mm}
\end{figure*}

\subsection{Animation in 3D Scenes}
\label{sec:animation}

The preceding stages produce a complete, animatable 3DGS avatar that can be driven by arbitrary SMPL poses via LBS with pose-dependent Gaussian maps (Sec.~\ref{sec:pose_dependent_maps}). Following AHA!~\cite{mir2025ahaanimatinghumanavatars}, we animate this avatar within reconstructed 3DGS scenes. To prevent out-of-distribution artifacts, we apply PCA-based pose projection at inference, following~\cite{li2024animatablegaussians}, constraining novel driving poses to the training distribution (details in the supplementary). Given a scene reconstructed as a static Gaussian set $\set{G}^{\Scn}$ from casual cell-phone video~\cite{kerbl2023gaussians} and a driving motion of $T''$ frames $\{\pose_t\}_{t=1}^{T''}$, we deform the avatar at each timestep and render $\mat{I}_{\mathrm{out}} = \rasterize\!\bigl(\deform(\set{G}^{\Hum}_t;\,\pose_t) \cup \set{G}^{\Scn};\;\camerapose\bigr)$, where $\set{G}^{\Hum}_t\!=\!S(\pose_t)$ uses the (PCA-projected) pose-dependent Gaussian maps, producing photorealistic novel-view videos of the reconstructed human animated within the target environment.

\section{Experiments}
\label{sec:experiments}

We evaluate AHOY along four axes: \emph{animation on YouTube} (Sec.~\ref{sec:comp_animatable}); \emph{static reconstruction on YouTube} (Sec.~\ref{sec:comp_static}); \emph{multi-view evaluation on BEHAVE} with ground-truth side views measuring novel-view and novel-pose quality under occlusion (Sec.~\ref{sec:comp_behave}); and \emph{component analysis} (Sec.~\ref{sec:ablation}). We further demonstrate animation within photorealistic 3DGS scenes (Sec.~\ref{sec:exp_scenes}).

\newcommand{\statw}{0.155\linewidth}
\newcommand{\stath}{2.8cm}
\newcommand{\stathocc}{2.4cm}
\begin{figure*}[t]
    \centering
    \setlength{\tabcolsep}{1pt}
    \renewcommand{\arraystretch}{0.6}
    \begin{tabular}{cccccc}
        {\tiny Input} &
        {\tiny \textbf{Ours}} &
        {\tiny SyncHuman~\cite{chen2025synchuman}} &
        {\tiny PSHuman~\cite{li2025pshuman}} &
        {\tiny LHM~\cite{qiu2025lhm}} &
        {\tiny IDOL~\cite{zhuang2025idol}} \\[2pt]
        \includegraphics[width=\statw,height=\stath,keepaspectratio]{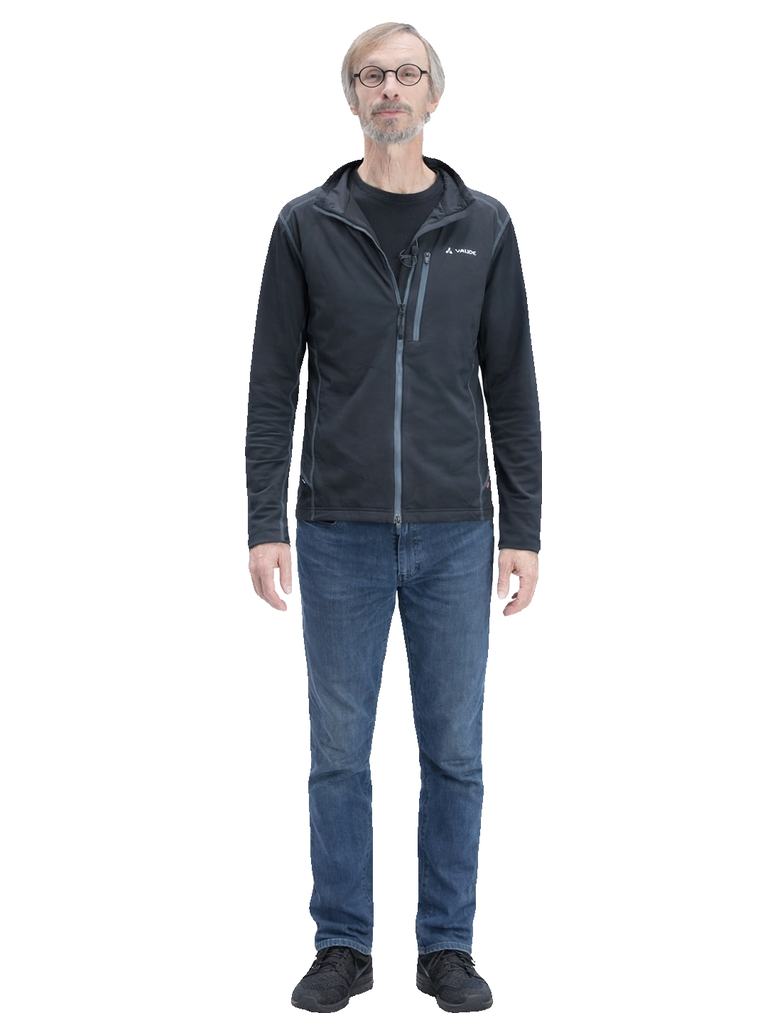} &
        \includegraphics[width=\statw,height=\stath,keepaspectratio]{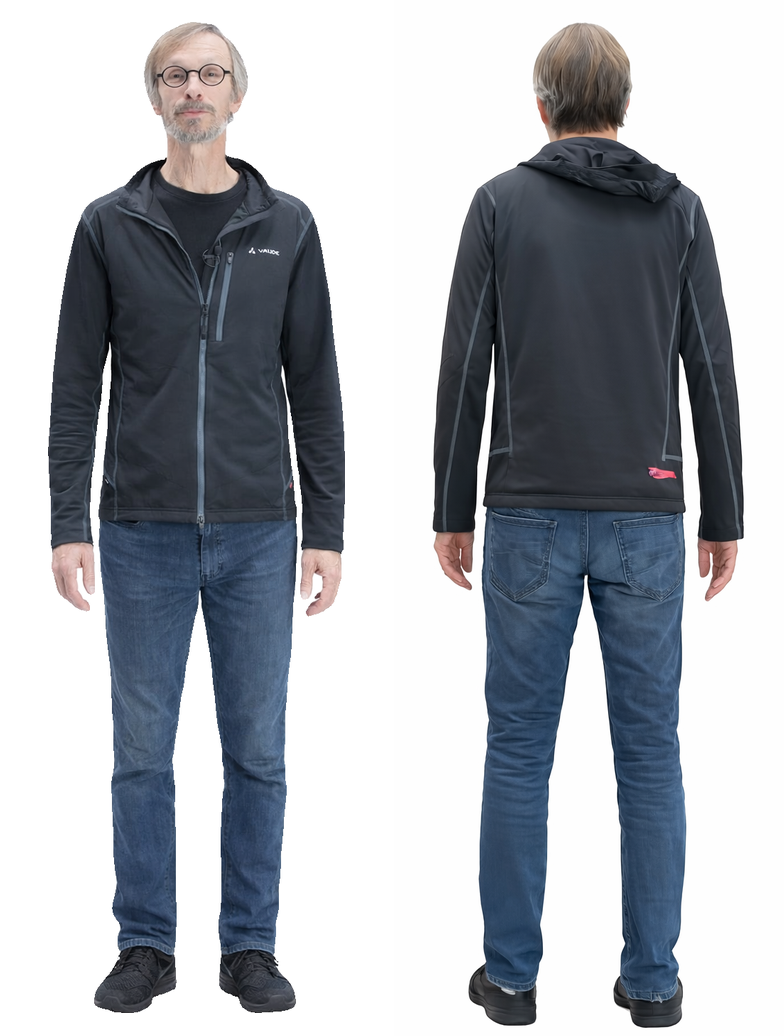} &
        \includegraphics[width=\statw,height=\stath,keepaspectratio]{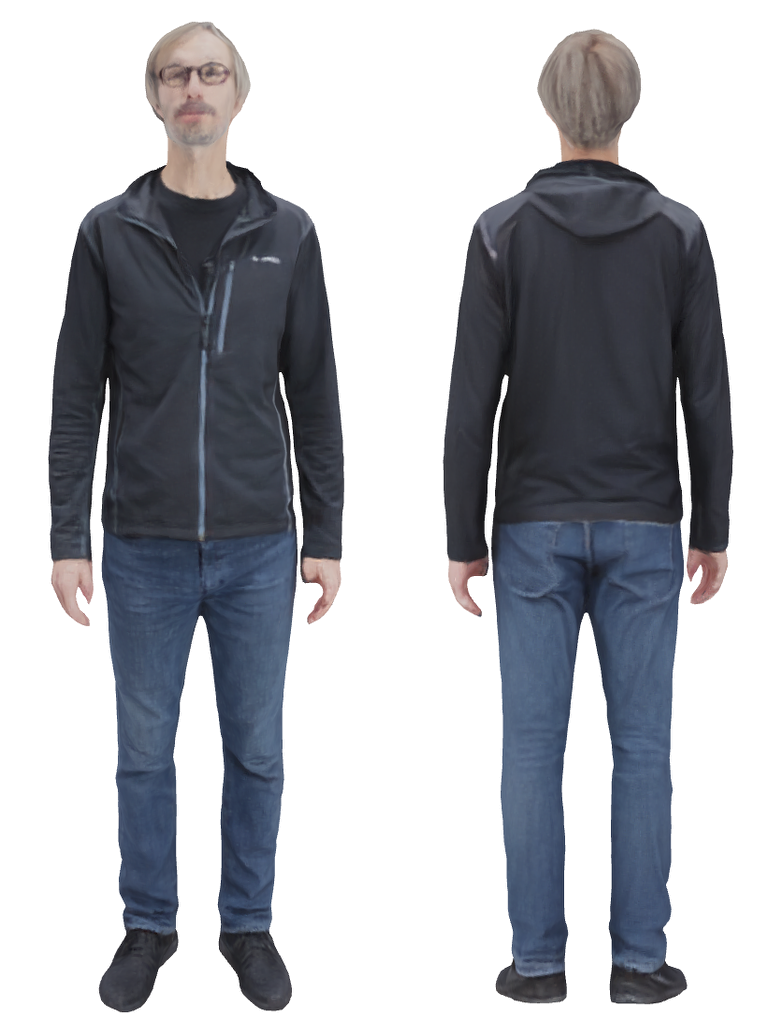} &
        \includegraphics[width=\statw,height=\stath,keepaspectratio]{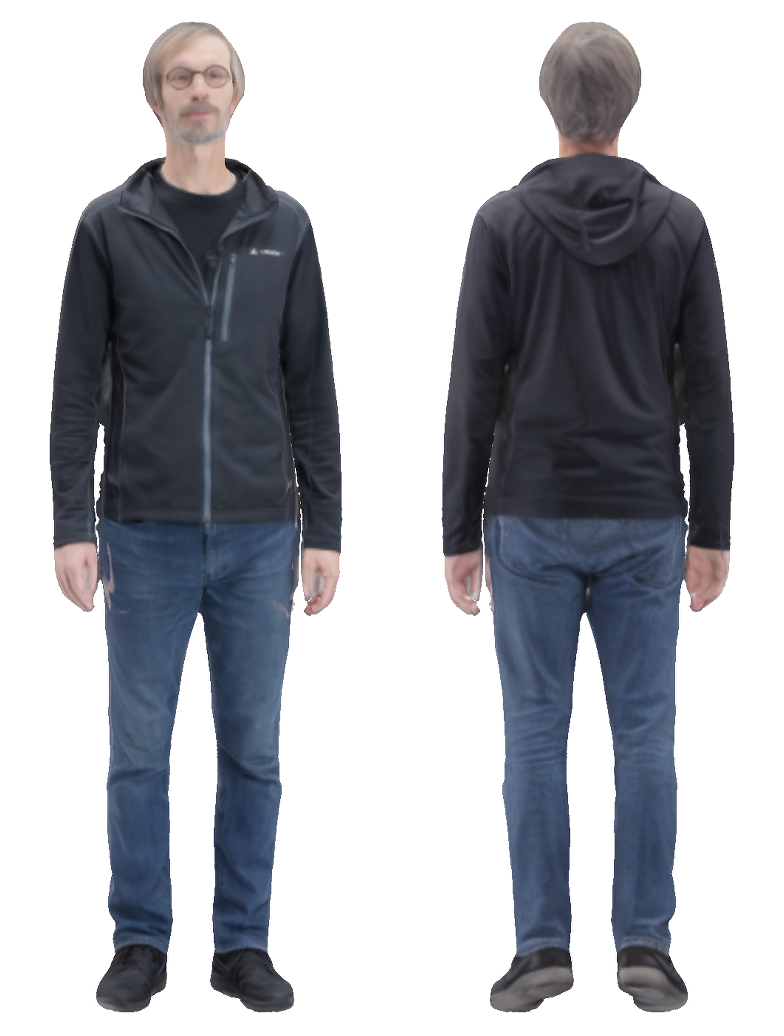} &
        \includegraphics[width=\statw,height=\stath,keepaspectratio]{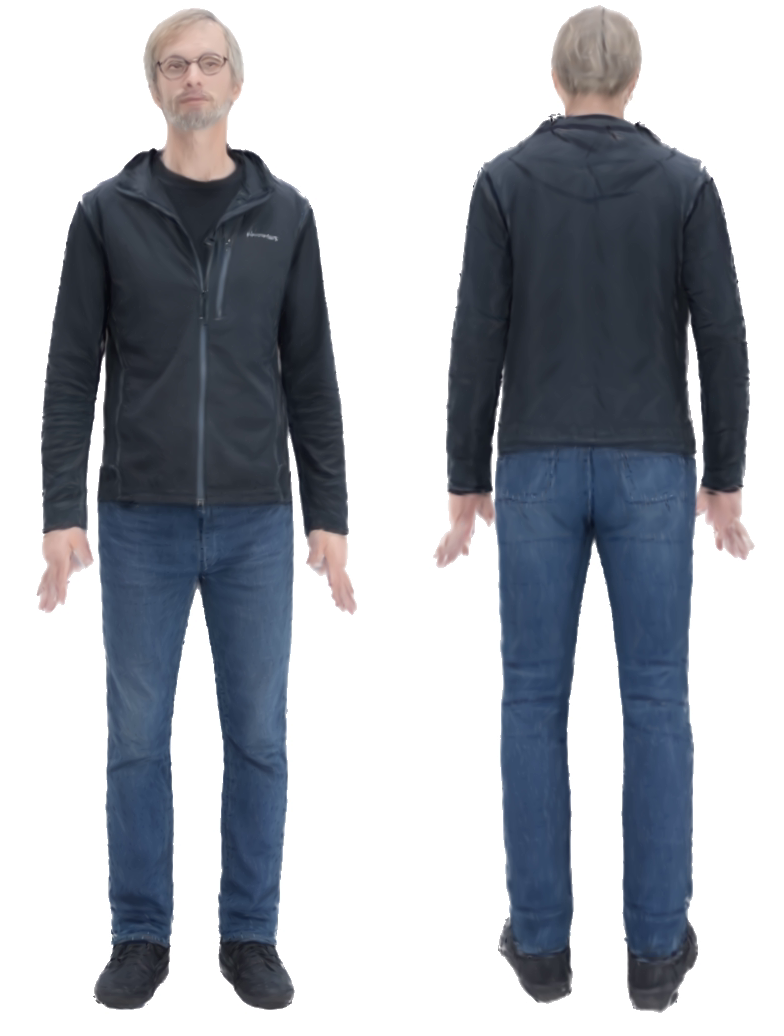} &
        \includegraphics[width=\statw,height=\stath,keepaspectratio]{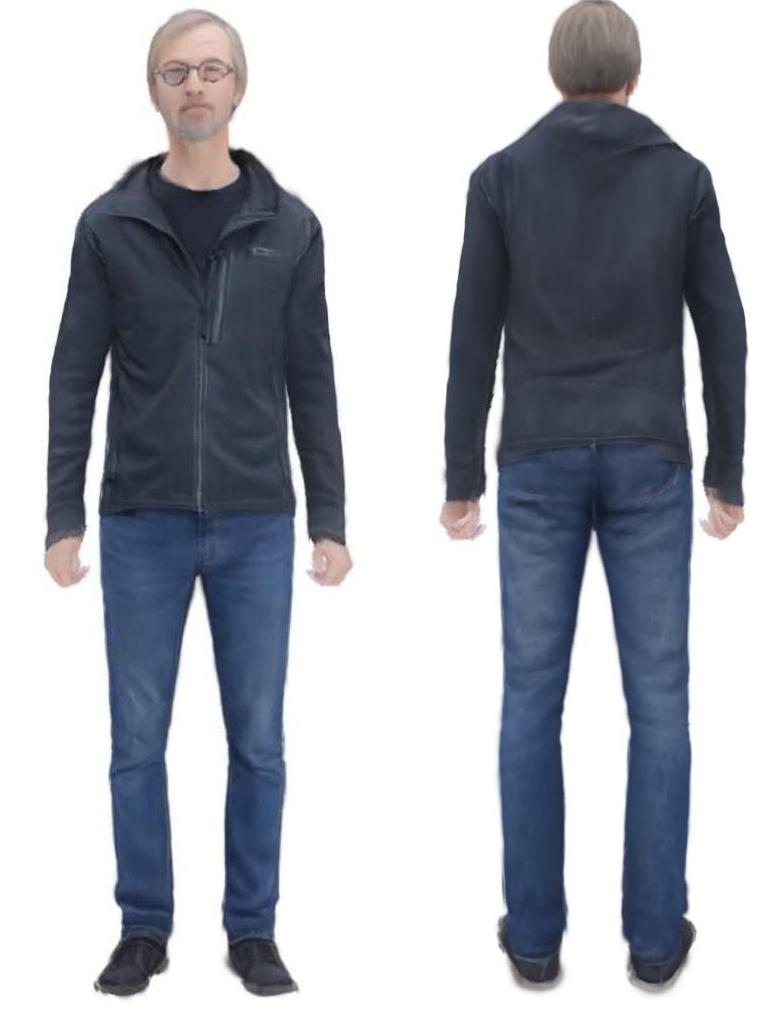} \\[-5pt]
        \multicolumn{6}{c}{\rule{0.95\linewidth}{0.4pt}} \\[-1pt]
        \includegraphics[width=\statw,height=\stathocc,keepaspectratio]{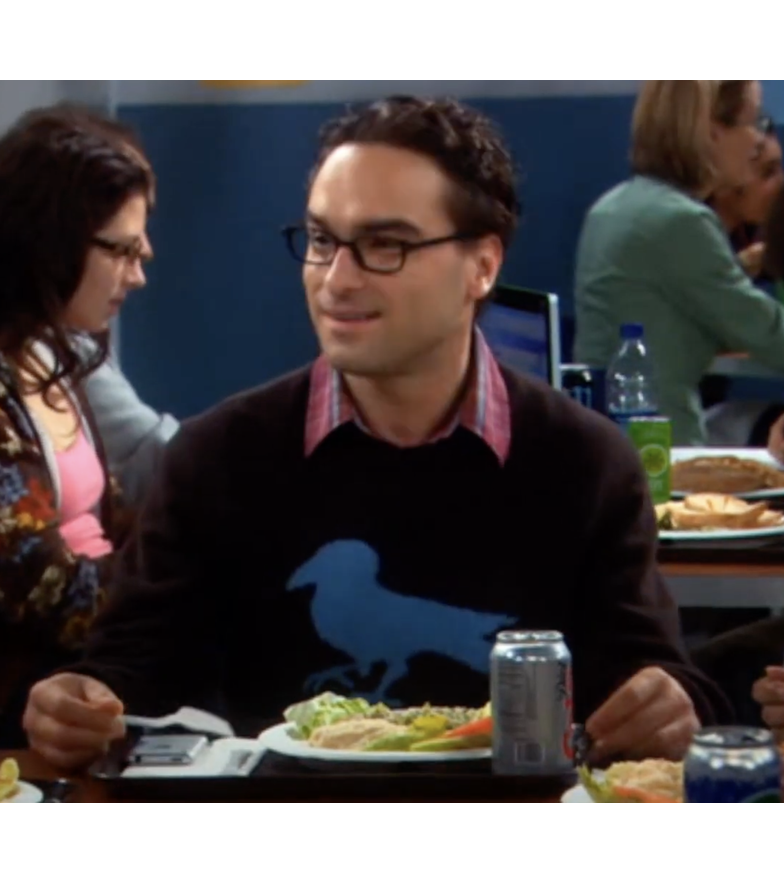} &
        \includegraphics[width=\statw,height=\stathocc,keepaspectratio]{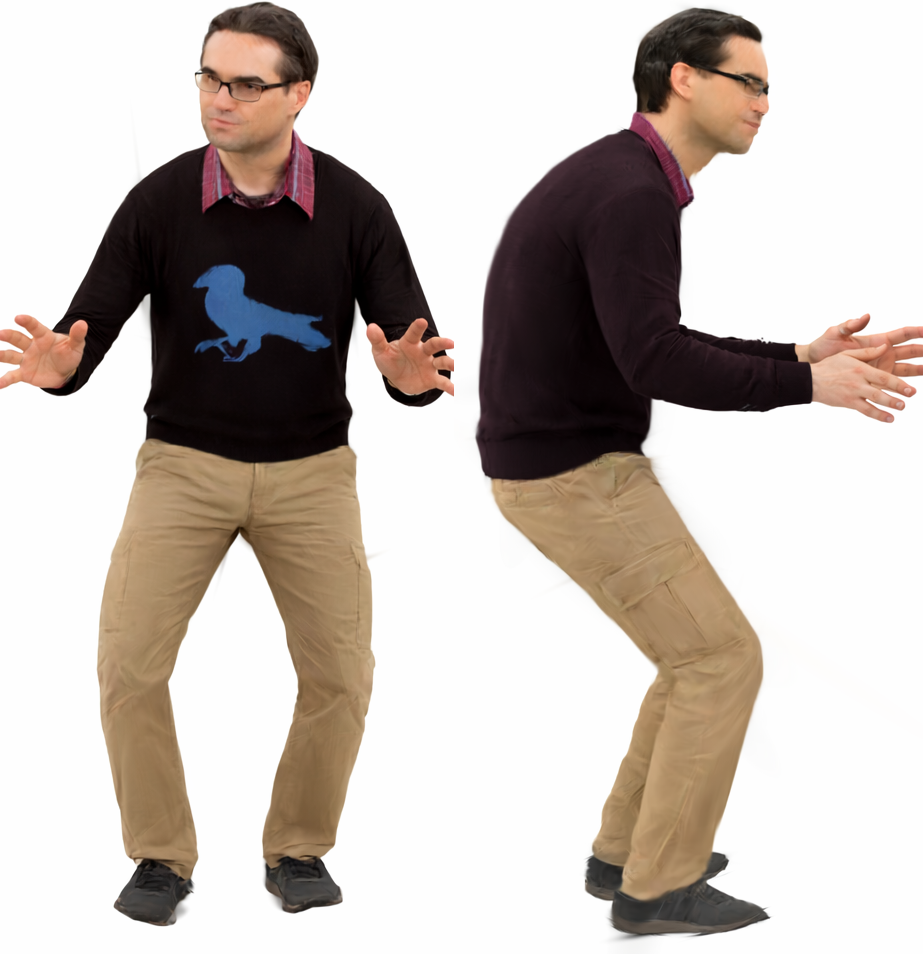} &
        \includegraphics[width=\statw,height=\stathocc,keepaspectratio]{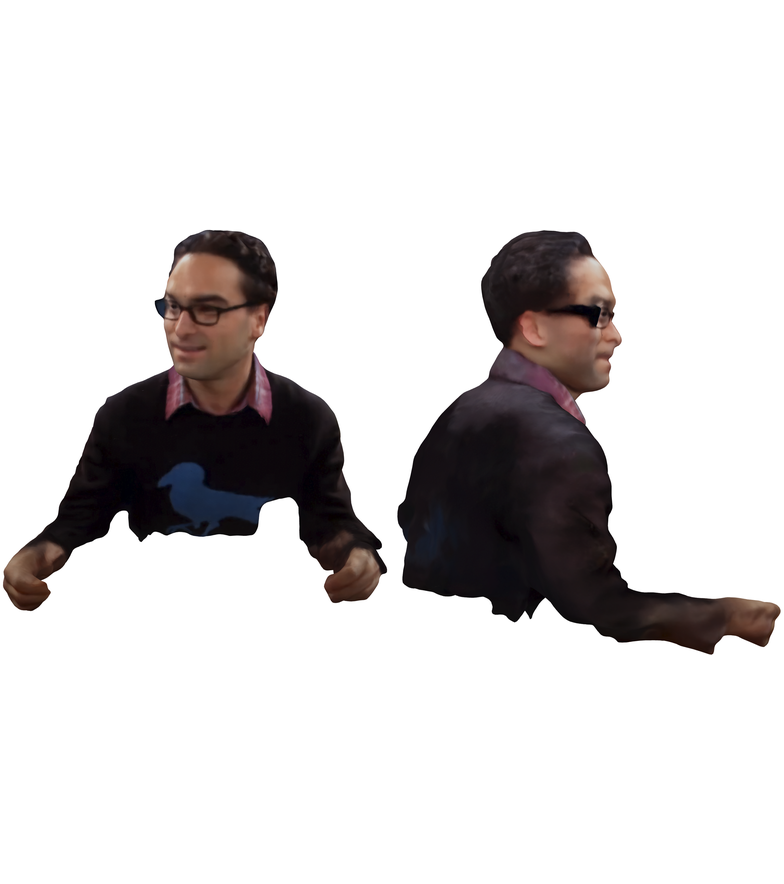} &
        \includegraphics[width=\statw,height=\stathocc,keepaspectratio]{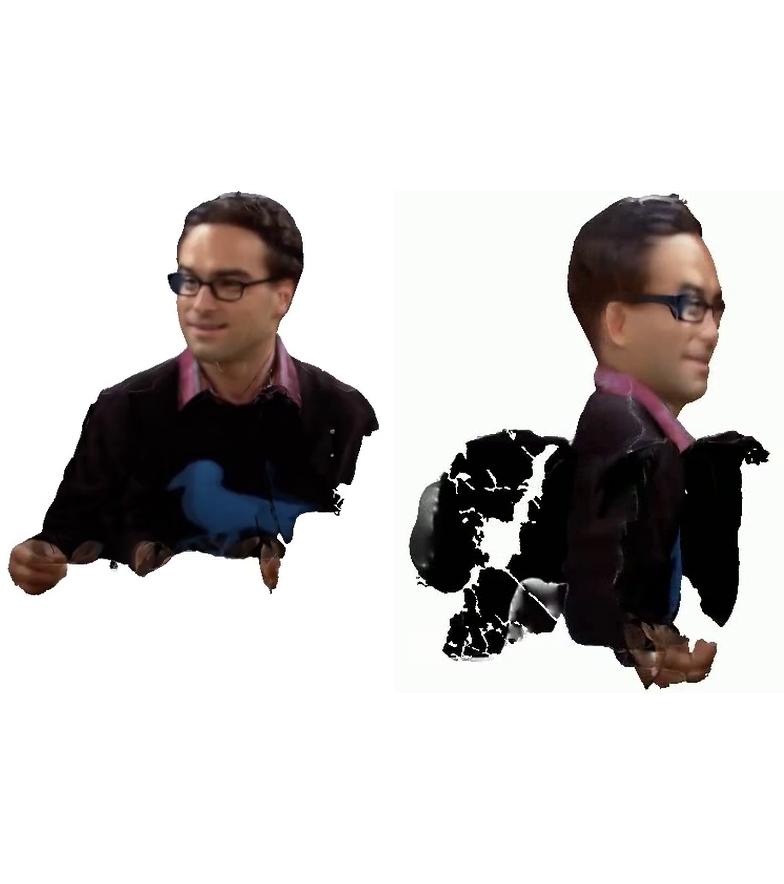} &
        \includegraphics[width=\statw,height=\stathocc,keepaspectratio]{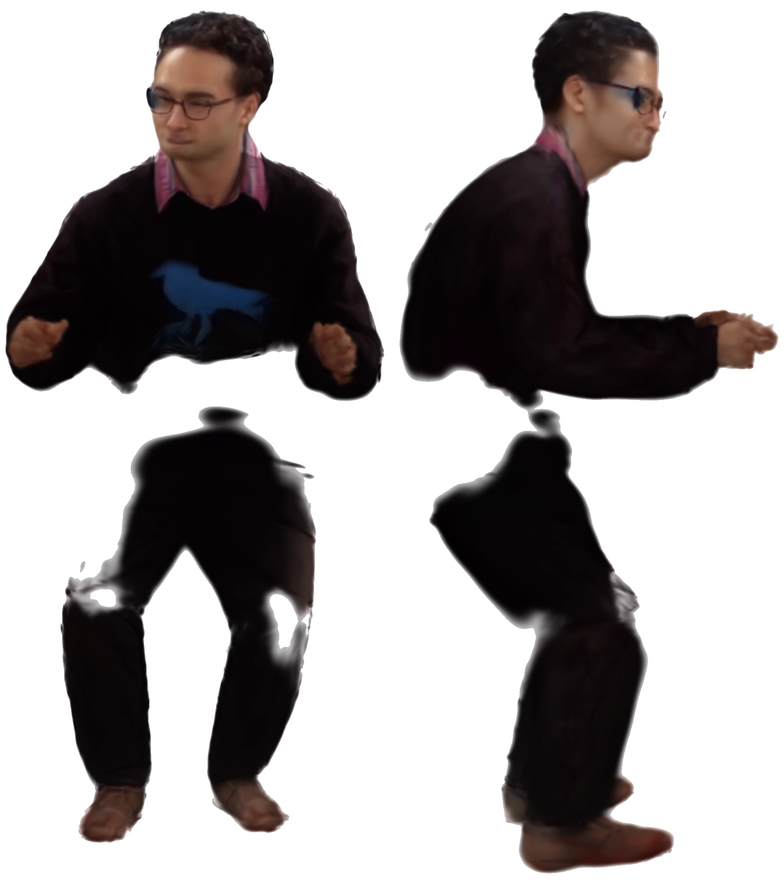} &
        \includegraphics[width=\statw,height=\stathocc,keepaspectratio]{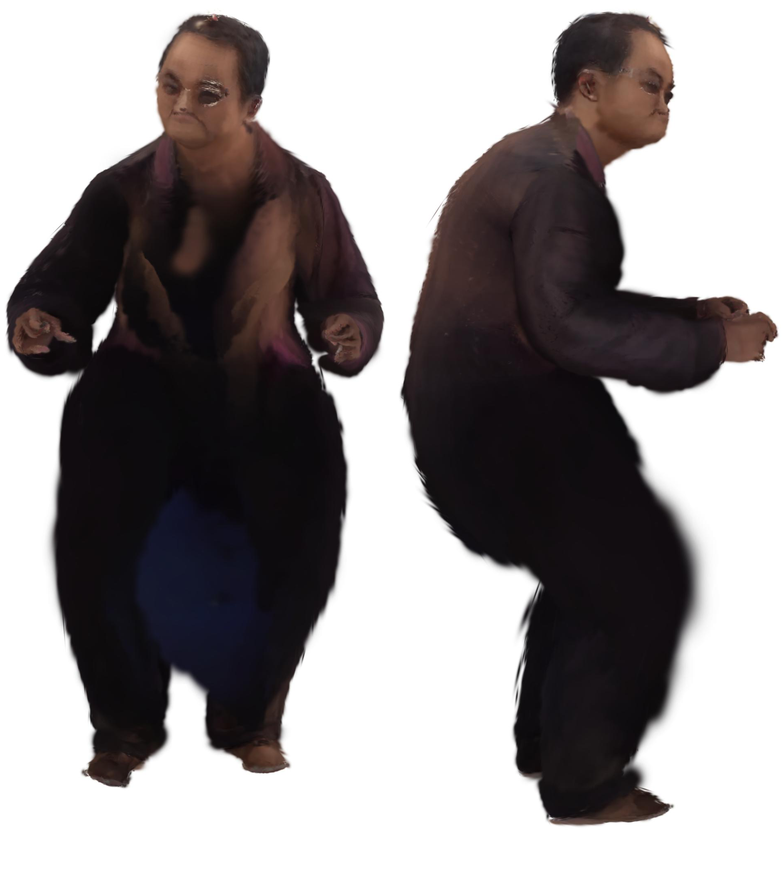} \\
    \end{tabular}
    \caption{\textbf{Static reconstruction comparison (Zoom In).} Top: canonical-pose input. Bottom: occluded input. Our animatable avatar is posed to match the target view.}
    \label{fig:comp_static}
\vspace{-2mm}
\end{figure*}

\begin{figure*}[t]
    \centering
    \vspace{8pt}
    \begin{overpic}[width=\linewidth]{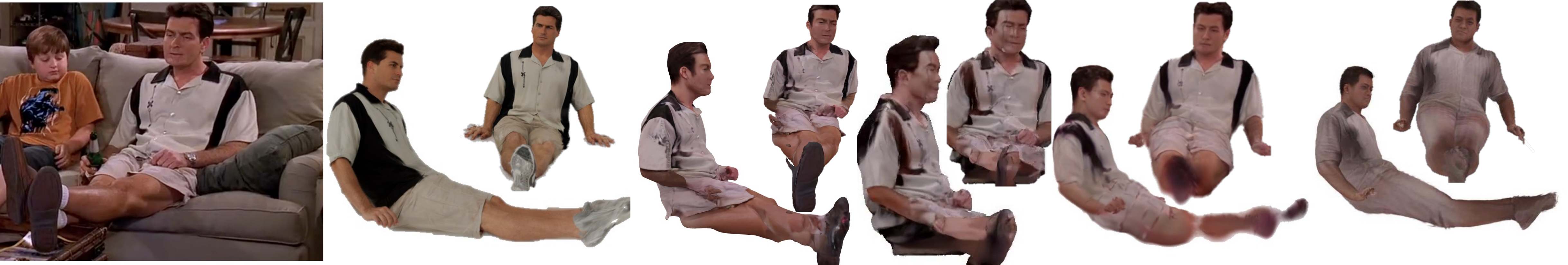}
        \put(7,18){\tiny Input}
        \put(28,18){\tiny \textbf{Ours}}
        \put(42,18){\tiny SyncHuman}
        \put(58,18){\tiny PSHuman}
        \put(73,18){\tiny LHM}
        \put(89,18){\tiny IDOL}
    \end{overpic}
    \caption{\textbf{Reconstruction under difficult fully visible poses.} Static methods struggle to reconstruct humans under difficult poses, while our method allows reposing of 3DGS avatars to match the target pose.}
    \label{fig:comp_static_sheen}
\vspace{-2mm}
\end{figure*}

\subsection{Experimental Setup}
\label{sec:setup}

\textbf{Datasets.}
We evaluate on two benchmarks that provide naturally occluded human subjects.
\textbf{BEHAVE}~\cite{bhatnagar22behave} contains multi-view captures (4 Kinect cameras) of humans interacting with objects, producing significant body occlusion. We evaluate on 8 sequences. The multi-view ground truth enables rigorous quantitative evaluation.
\textbf{YouTube videos.} We collect 50 videos of humans in diverse everyday activities with significant occlusion. For these sequences, ground-truth multi-view images are unavailable, so we compare against held-out frames from the original video. \textbf{Metrics.}
We report PSNR, SSIM~\cite{wang2004ssim}, and LPIPS~\cite{zhang2018lpips}.

\textbf{Baselines.}
Our method is video-based, while all baselines are single-image methods. We consider recent animatable avatar methods with public code: LHM~\cite{qiu2025lhm} and IDOL~\cite{zhuang2025idol}. We evaluate baselines under two input settings: (1)~receiving the original occluded frame (first video frame) directly, and (2)~receiving the unoccluded canonical-pose image generated by our pipeline, giving them the best possible input. We additionally compare with static reconstruction methods PSHuman~\cite{li2025pshuman} and SyncHuman~\cite{chen2025synchuman}, which do not produce animatable avatars. We note that AdaHuman~\cite{huang2025adahuman}, Dream-Lift-Animate~\cite{buehler2026dreamliftanimate}, MoGA~\cite{dong2025moga}, and MEAT~\cite{wang2025meat} are plausible baselines, but none have publicly available code.

\subsection{Animation on YouTube}
\label{sec:comp_animatable}

We compare with LHM~\cite{qiu2025lhm} and IDOL~\cite{zhuang2025idol} (the only baselines that produce animatable avatars) under two input settings (Tab.~\ref{tab:youtube_anim}, Fig.~\ref{fig:comp_anim}). We animate reconstructed avatars to novel poses and compare against held-out video frames. \textbf{Canonical-pose input.}
To provide a fairer comparison, we give baselines the unoccluded canonical-pose images generated by our pipeline (Sec.~\ref{sec:texture_mapping}). Even with complete input, our full pipeline outperforms these methods (Fig.~\ref{fig:comp_anim}, top).
\textbf{Occluded input.}
When baselines receive the original occluded video frames, the realism of their avatars drops significantly (Fig.~\ref{fig:comp_anim}, bottom).

\subsection{Static Reconstruction on YouTube}
\label{sec:comp_static}

\definecolor{best}{RGB}{220,250,220}
\definecolor{second}{RGB}{240,245,220}
\definecolor{headerbg}{RGB}{200,220,240}

\begin{table*}[t]
\tiny
\setlength{\tabcolsep}{1.5pt}
\renewcommand{\arraystretch}{1.15}
\centering
\caption{\textbf{Quantitative evaluation.} Left: static reconstruction on YouTube videos under occluded and canonical-pose input. Right: evaluation on BEHAVE with novel-view and novel-pose settings.}
\label{tab:youtube_static}\label{tab:behave}
\begin{tabular}{l|ccc|ccc|ccc|ccc}
\toprule
\rowcolor{headerbg}
& \multicolumn{6}{c|}{\textbf{YouTube Static}} & \multicolumn{6}{c}{\textbf{BEHAVE}} \\
\rowcolor{headerbg}
& \multicolumn{3}{c|}{\textbf{Occluded}} & \multicolumn{3}{c|}{\textbf{Canonical}} & \multicolumn{3}{c|}{\textbf{Novel View}} & \multicolumn{3}{c}{\textbf{Novel Pose}} \\
\rowcolor{headerbg}
\textbf{Method} & PSNR$\uparrow$ & SSIM$\uparrow$ & LPIPS$\downarrow$ & PSNR$\uparrow$ & SSIM$\uparrow$ & LPIPS$\downarrow$ & PSNR$\uparrow$ & SSIM$\uparrow$ & LPIPS$\downarrow$ & PSNR$\uparrow$ & SSIM$\uparrow$ & LPIPS$\downarrow$ \\
\midrule
PSHuman~\cite{li2025pshuman} & 19.47 & 0.811 & 0.197 & 21.88 & 0.858 & 0.145 & 18.47 & 0.812 & 0.198 & \multicolumn{3}{c}{\textit{n/a}} \\
SyncHuman~\cite{chen2025synchuman} & 19.82 & 0.819 & 0.189 & 22.03 & 0.863 & 0.140 & 19.34 & 0.831 & 0.172 & \multicolumn{3}{c}{\textit{n/a}} \\
LHM~\cite{qiu2025lhm}       & 19.12 & 0.803 & 0.207 & 21.18 & 0.845 & 0.160 & 18.21 & 0.807 & 0.203 & 16.93 & 0.774 & 0.238 \\
IDOL~\cite{zhuang2025idol}   & 17.31 & 0.771 & 0.241 & 19.78 & 0.819 & 0.189 & 17.38 & 0.783 & 0.231 & 15.87 & 0.748 & 0.267 \\
\midrule
\rowcolor{best}
\textbf{Ours}          & \textbf{22.01} & \textbf{0.881} & \textbf{0.109} & \textbf{23.05} & \textbf{0.892} & \textbf{0.098} & \textbf{24.12} & \textbf{0.905} & \textbf{0.090} & \textbf{22.81} & \textbf{0.889} & \textbf{0.110} \\
\bottomrule
\end{tabular}
\vspace{-2mm}
\end{table*}

\begin{table*}[t]
\tiny
\setlength{\tabcolsep}{3pt}
\renewcommand{\arraystretch}{1.15}
\centering
\caption{\textbf{Animation comparison on YouTube videos.} We animate reconstructed avatars to novel poses and compare against held-out video frames.}
\label{tab:youtube_anim}
\begin{tabular}{l|ccc|ccc}
\toprule
\rowcolor{headerbg}
& \multicolumn{3}{c|}{\textbf{Occluded input}} & \multicolumn{3}{c}{\textbf{Canonical-pose input}} \\
\rowcolor{headerbg}
\textbf{Method} & \textbf{PSNR}$\uparrow$ & \textbf{SSIM}$\uparrow$ & \textbf{LPIPS}$\downarrow$ & \textbf{PSNR}$\uparrow$ & \textbf{SSIM}$\uparrow$ & \textbf{LPIPS}$\downarrow$ \\
\midrule
LHM~\cite{qiu2025lhm}       & 19.03 & 0.821 & 0.181 & 20.17 & 0.842 & 0.159 \\
IDOL~\cite{zhuang2025idol}   & 16.52 & 0.769 & 0.231 & 18.52 & 0.808 & 0.191 \\
\midrule
\rowcolor{best}
\textbf{AHOY (Ours)}          & \textbf{22.81} & \textbf{0.887} & \textbf{0.107} & \textbf{22.83} & \textbf{0.888} & \textbf{0.106} \\
\bottomrule
\end{tabular}
\vspace{-2mm}
\end{table*}

We compare with PSHuman~\cite{li2025pshuman}, SyncHuman~\cite{chen2025synchuman}, LHM~\cite{qiu2025lhm}, and IDOL~\cite{zhuang2025idol} on YouTube videos under two input settings (Tab.~\ref{tab:youtube_static}, Figs.~\ref{fig:comp_static}--\ref{fig:comp_static_sheen}). Each method reconstructs from its respective input, and all reconstructions are rendered from the input viewpoint and compared against the original video frame. \textbf{Occluded input.} Under occlusion, all baselines fail to produce complete reconstructions, while our method generates a full avatar that can be accurately posed to match the target view.
\textbf{Canonical-pose input.} To provide a fairer comparison, we give all baselines the unoccluded canonical-pose images generated by our pipeline. Even with complete input, our method outperforms all baselines.
In Fig.~\ref{fig:comp_static_sheen}, we demonstrate that static methods struggle to reconstruct humans under difficult poses, while our method allows reposing of avatars to match the target pose.

\subsection{BEHAVE Evaluation}
\label{sec:comp_behave}

\newcommand{\behw}{0.078\linewidth}
\newcommand{\behh}{2.8cm}
\begin{figure*}[t]
    \centering
    \setlength{\tabcolsep}{0.5pt}
    \renewcommand{\arraystretch}{0.6}
    \begin{tabular}{cc cc cc cc cc cc}
        {\tiny Input} & {\tiny GT} &
        \multicolumn{2}{c}{\tiny \textbf{Ours}} &
        \multicolumn{2}{c}{\tiny SyncHuman~\cite{chen2025synchuman}} &
        \multicolumn{2}{c}{\tiny PSHuman~\cite{li2025pshuman}} &
        \multicolumn{2}{c}{\tiny LHM~\cite{qiu2025lhm}} &
        \multicolumn{2}{c}{\tiny IDOL~\cite{zhuang2025idol}} \\[2pt]
        \includegraphics[width=\behw,height=\behh,keepaspectratio]{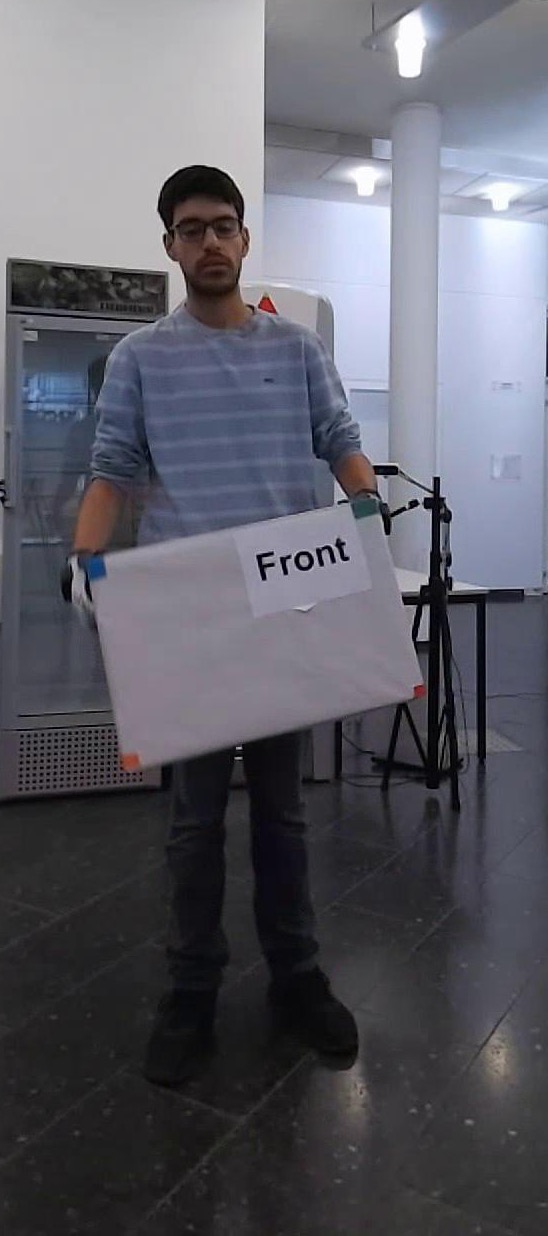} &
        \includegraphics[width=\behw,height=\behh,keepaspectratio]{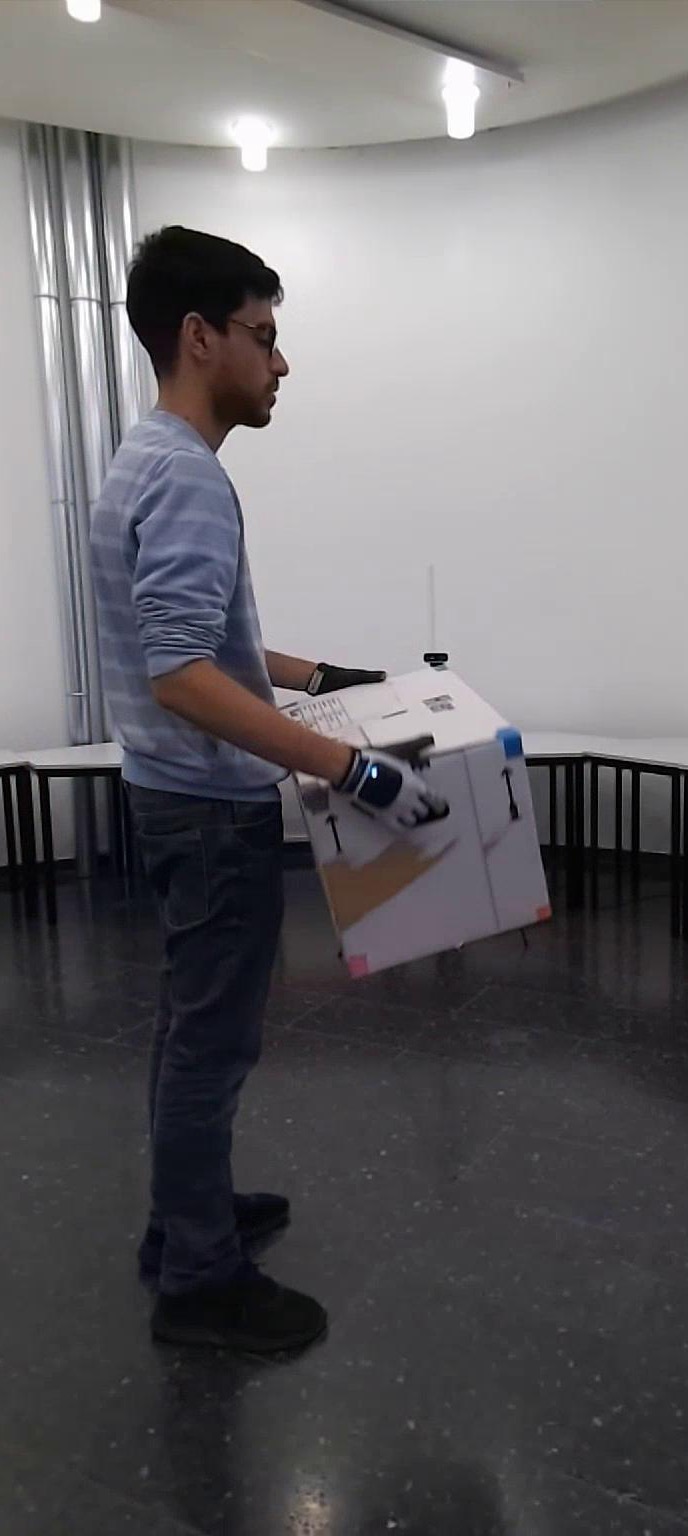} &
        \includegraphics[width=\behw,height=\behh,keepaspectratio]{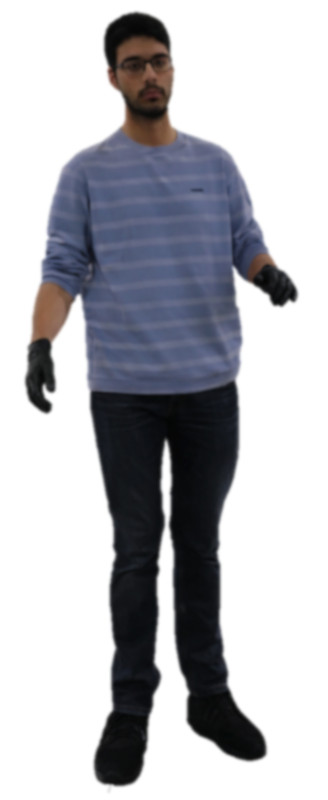} &
        \includegraphics[width=\behw,height=\behh,keepaspectratio]{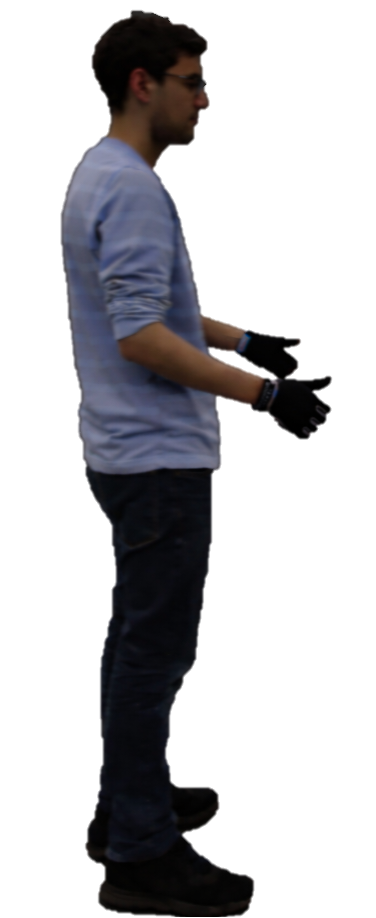} &
        \includegraphics[width=\behw,height=\behh,keepaspectratio]{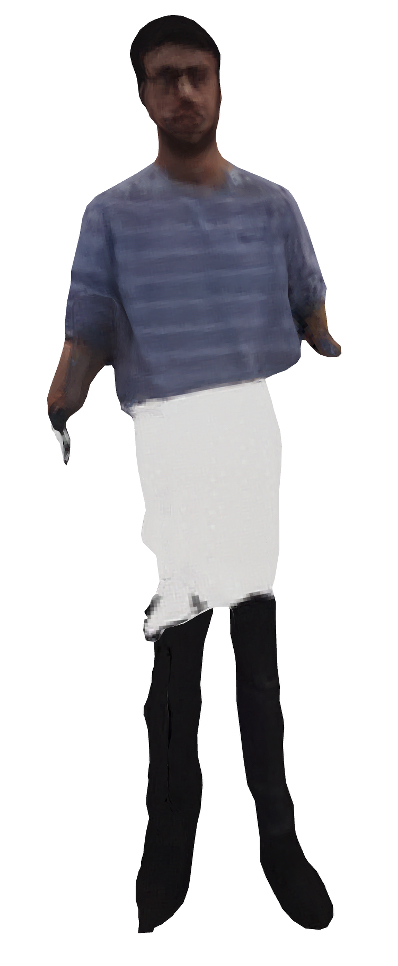} &
        \includegraphics[width=\behw,height=\behh,keepaspectratio]{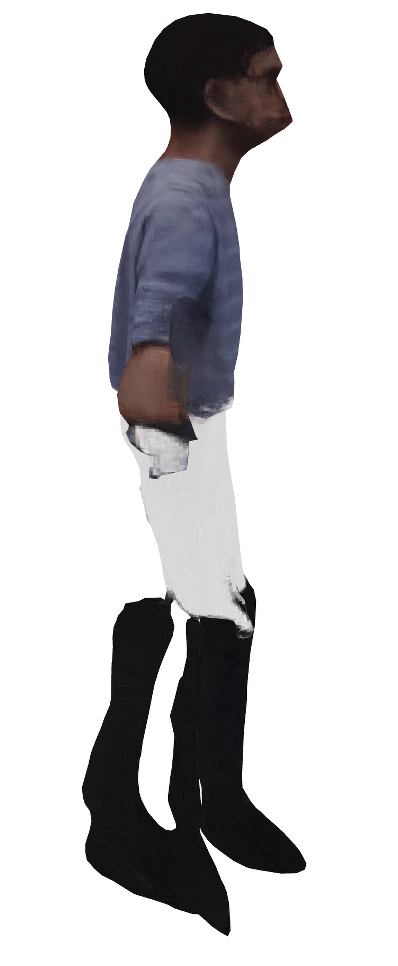} &
        \includegraphics[width=\behw,height=\behh,keepaspectratio]{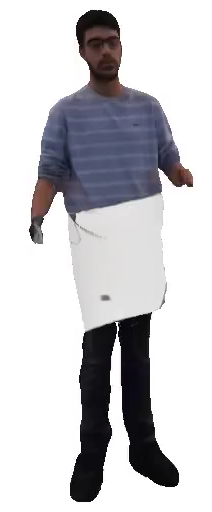} &
        \includegraphics[width=\behw,height=\behh,keepaspectratio]{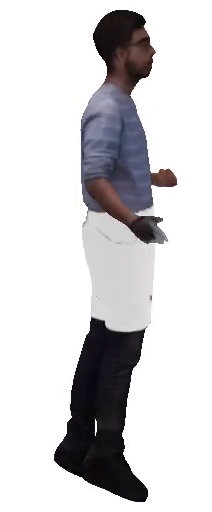} &
        \includegraphics[width=\behw,height=\behh,keepaspectratio]{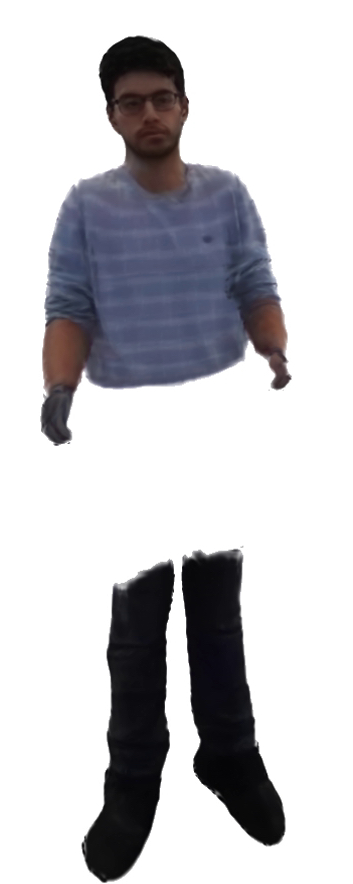} &
        \includegraphics[width=\behw,height=\behh,keepaspectratio]{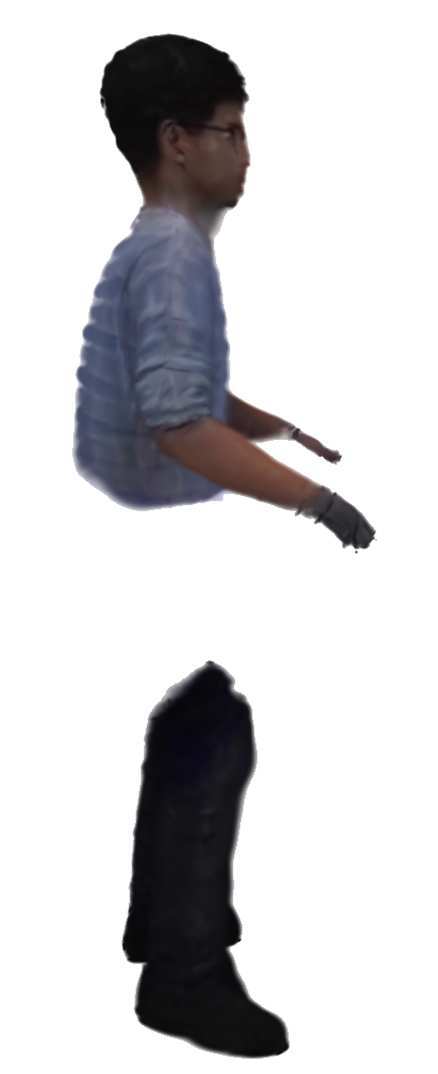} &
        \includegraphics[width=\behw,height=\behh,keepaspectratio]{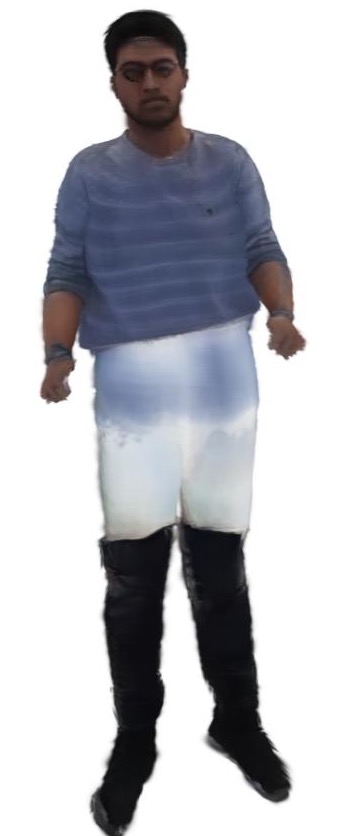} &
        \includegraphics[width=\behw,height=\behh,keepaspectratio]{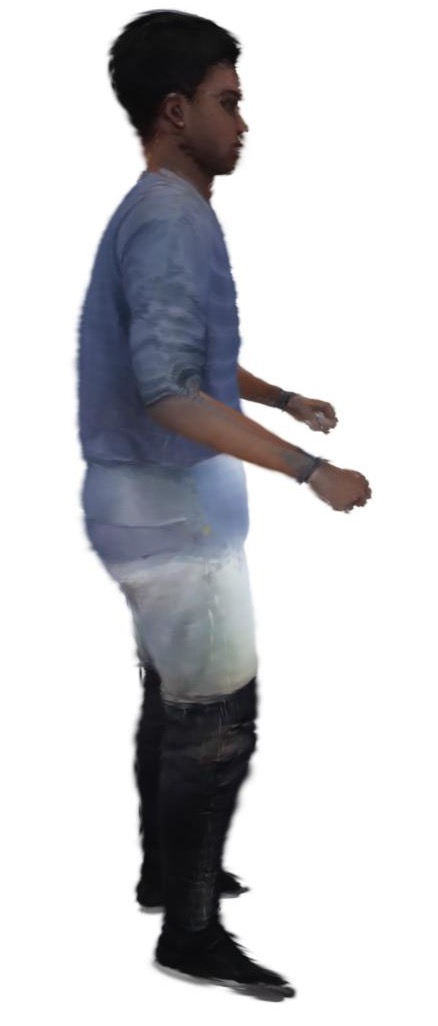} \\
    \end{tabular}
    \caption{\textbf{Novel-view evaluation on BEHAVE.} We render each reconstruction from held-out BEHAVE side cameras and compare against the ground-truth side views.}
    \label{fig:behave_novelview}
\vspace{-1mm}
\end{figure*}
\begin{table*}[t]
\tiny
\setlength{\tabcolsep}{3pt}
\renewcommand{\arraystretch}{1.15}
\centering
\caption{\textbf{Ablation study on BEHAVE.} Each row removes one component from the full pipeline. \emph{Novel View}: render from held-out side cameras. \emph{Novel Pose}: animate to unseen ground-truth SMPL-X poses and render from held-out side cameras.}
\label{tab:ablation}
\begin{tabular}{l|ccc|ccc}
\toprule
\rowcolor{headerbg}
& \multicolumn{3}{c|}{\textbf{Novel View}} & \multicolumn{3}{c}{\textbf{Novel Pose}} \\
\rowcolor{headerbg}
\textbf{Configuration} & \textbf{PSNR}$\uparrow$ & \textbf{SSIM}$\uparrow$ & \textbf{LPIPS}$\downarrow$ & \textbf{PSNR}$\uparrow$ & \textbf{SSIM}$\uparrow$ & \textbf{LPIPS}$\downarrow$ \\
\midrule
\rowcolor{best}
\textbf{AHOY (full)}                        & \textbf{24.12} & \textbf{0.905} & \textbf{0.090} & \textbf{22.81} & \textbf{0.889} & \textbf{0.110} \\
\midrule
(A) Coarse avatar only                      & 16.10 & 0.753 & 0.258 & 14.30 & 0.718 & 0.301 \\
(B) w/o RF-Inversion                        & 21.10 & 0.862 & 0.143 & 19.40 & 0.838 & 0.168 \\
(C) w/o map/LBS decoupling                  & 22.60 & 0.886 & 0.118 & 20.60 & 0.856 & 0.152 \\
(D) w/o head/body split                     & 23.60 & 0.898 & 0.098 & 22.20 & 0.882 & 0.120 \\
\bottomrule
\end{tabular}
\vspace{-2mm}
\end{table*}

We evaluate all methods on BEHAVE (Tab.~\ref{tab:behave}, Fig.~\ref{fig:behave_novelview}). All methods receive the occluded front-camera view as input. For \emph{novel-view} evaluation, we render each reconstruction from held-out side cameras and compare against ground-truth side views. For \emph{novel-pose} evaluation, we animate the reconstructed avatars to unseen ground-truth SMPL-X poses and render from held-out side cameras. Our method significantly outperforms all baselines in both settings.

\subsection{Ablation Study}
\label{sec:ablation}
\begin{figure*}[t]
    \centering
    \begin{overpic}[width=\linewidth,percent]{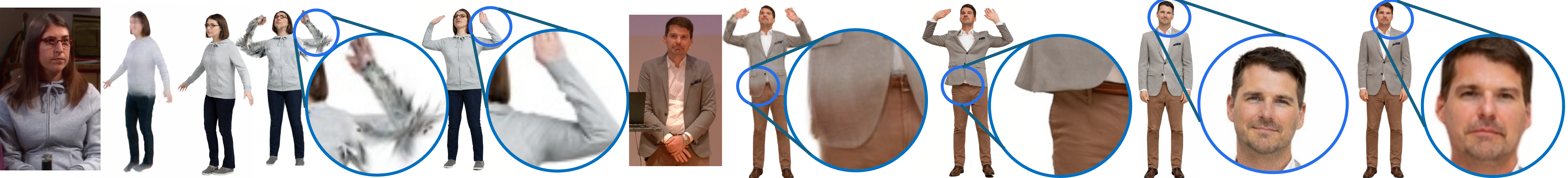}
        \put(-3,-4){\parbox{3em}{\centering\tiny Input}}
        \put(0,-4){\parbox{8em}{\centering\tiny (A) Coarse Full}}
        \put(11,-4){\parbox{12em}{\centering\tiny (B) w/o RF-Inv\;\;Full}}
        \put(38,-4){\parbox{3em}{\centering\tiny Input}}
        \put(40,-4){\parbox{12em}{\centering\tiny (C) w/o Pose Maps\;\;Full}}
        \put(68,-4){\parbox{10em}{\centering\tiny (D) w/o Identity \;\;Full}}
    \end{overpic}
    \vspace{1.5mm}
    \caption{\textbf{Visual ablations.} Each pair shows the ablated result (w/o) and the full pipeline. (A)~Without hallucination, the coarse-only avatar lacks detail in unobserved regions. (B)~Without RF-Inversion, hallucinated supervision retains rendering artifacts. (C)~Without map/LBS decoupling, multi-view inconsistencies cause blurring. (D)~Without the head/body split, facial identity is lost.}
    \label{fig:ablation}
\end{figure*}

We ablate each key component of our pipeline on BEHAVE to isolate its contribution, evaluating both novel-view and novel-pose settings as in Sec.~\ref{sec:comp_behave} (Tab.~\ref{tab:ablation},
Fig.~\ref{fig:ablation}). \textbf{(A) Coarse avatar only.} We skip the hallucination and full avatar stages, using only the coarse canonical-map avatar (Sec.~\ref{sec:coarse_avatar}). This removes pose-dependent appearance and the diffusion-based supervision for occluded regions.
\textbf{(B) Without RF-Inversion.}
We skip RF-Inversion and supervise the avatar without any new hallucinated videos of the person in new poses; this results in significant artifacts in unobserved regions (Sec.~\ref{sec:rf_inversion}).
\textbf{(C) Without map-pose/LBS-pose decoupling.}
We do not use pose-dependent maps for StyleUNet training and do not optimize LBS deformation, removing the decoupling that absorbs multi-view inconsistencies (Sec.~\ref{sec:joint_optimization}).
\textbf{(D) Without head/body split.}
We supervise the entire avatar, including the head, using the hallucinated Wan videos, without the separate head path (Sec.~\ref{sec:head_identity}).

\begin{figure*}[t]
    \centering
    \setlength{\tabcolsep}{2pt}
    \renewcommand{\arraystretch}{0.6}
    \begin{tabular}{cc}
        \includegraphics[width=0.49\linewidth,height=3.5cm,keepaspectratio]{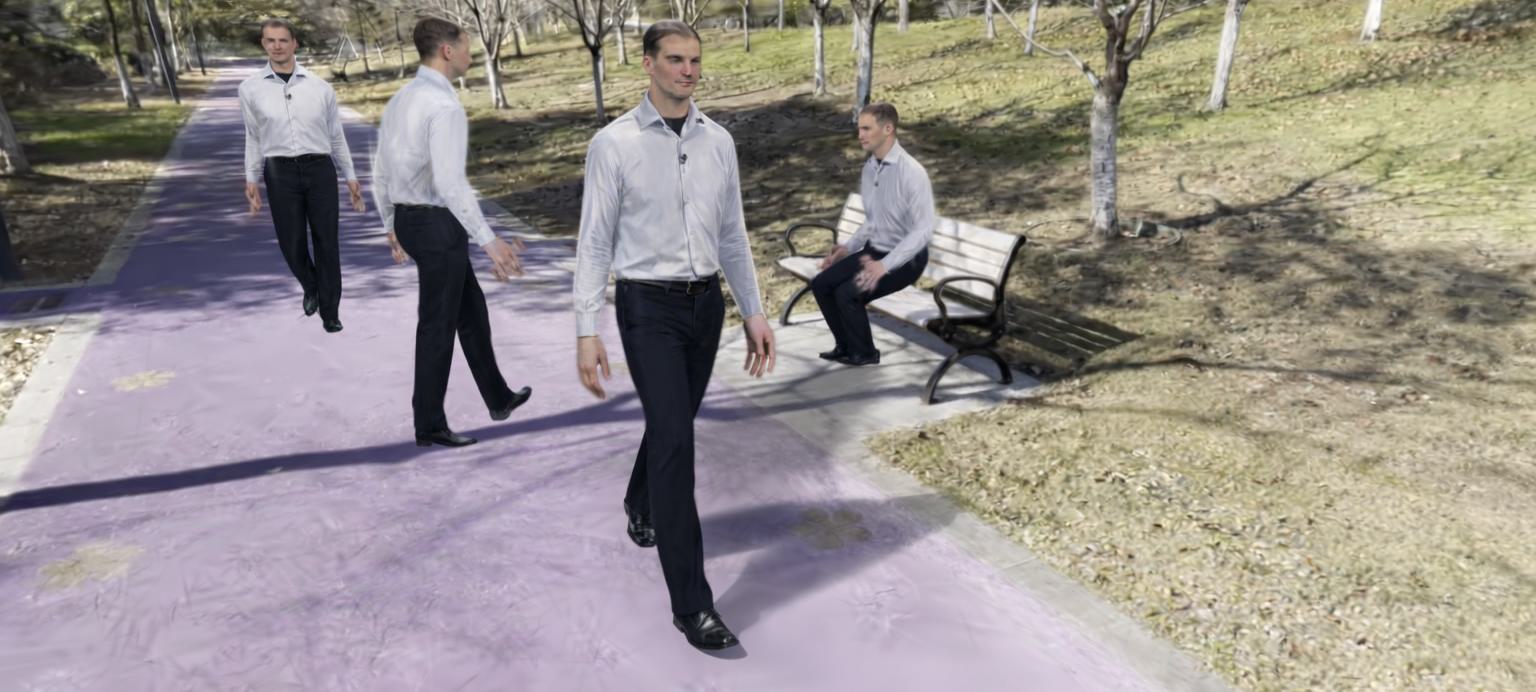} &
        \includegraphics[width=0.49\linewidth,height=3.5cm,keepaspectratio]{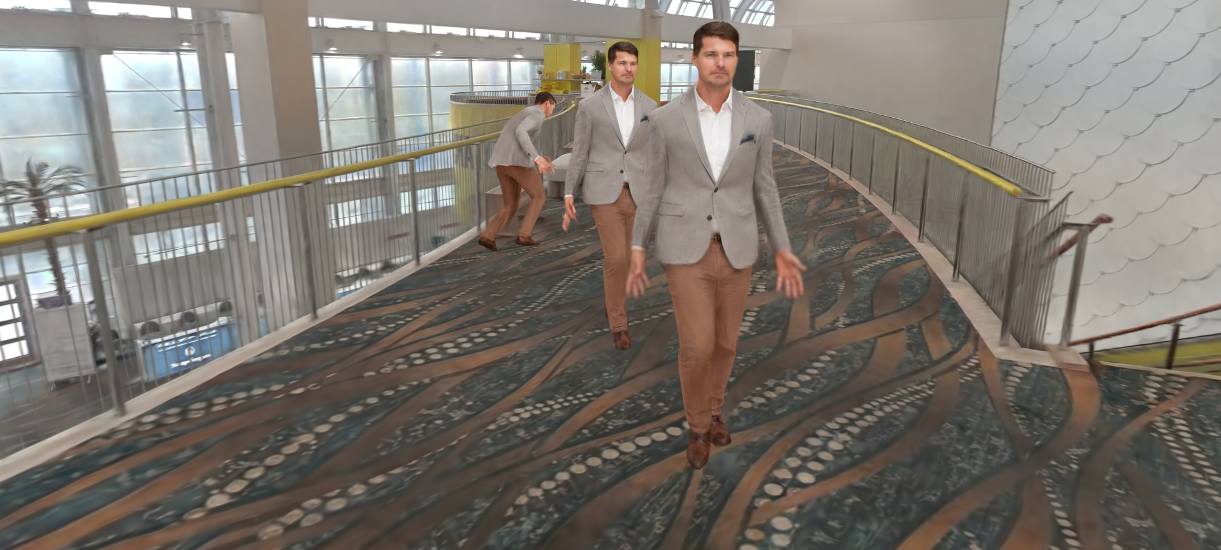} \\
    \end{tabular}
    \caption{\textbf{Animation in 3DGS scenes.} Avatars from occluded YouTube video, animated and composited into 3DGS scenes from casual cell-phone video.}
    \label{fig:scenes}
\vspace{-6mm}
\end{figure*}

\subsection{Animation in 3D Scenes}
\label{sec:exp_scenes}

We demonstrate that the avatars reconstructed by AHOY are robust enough to be animated with novel poses and composited into 3DGS scenes reconstructed from casual cell-phone video (Fig.~\ref{fig:scenes}). Please see the supplementary for videos.

\section{Conclusion}
\label{sec:conclusion}

We have presented AHOY, a method that reconstructs complete, animatable 3D Gaussian avatars from in-the-wild monocular video despite heavy occlusion, unlocking YouTube-scale footage as a source of 3D human assets. Our key insight is to use an identity-finetuned video diffusion model as a bridge between imperfect coarse renderings and dense, realistic supervision: by embedding coarse avatar renderings into the model's latent space via RF-Inversion and decoding them back, the identity prior fills in previously unobserved regions while preserving body layout. The resulting avatars are robust enough to be animated with novel poses and composited into 3DGS scenes reconstructed from casual cell-phone video.
Our method inherits limitations from its diffusion backbone: the identity-finetuned model may hallucinate plausible but incorrect details for body regions that are never observed, and the quality of the final avatar is bounded by the fidelity of the video diffusion model. Additionally, our pipeline requires several stages of optimization and diffusion inference, making it substantially slower than feed-forward methods.
Addressing these limitations suggests several promising directions: scaling to stronger video diffusion models as they improve and incorporating geometric consistency constraints into the hallucination stage. We hope that AHOY demonstrates that occlusion, long treated as a nuisance to be avoided, can be overcome, opening the door to learning from the vast diversity of human appearance captured in everyday video.

\paragraph{\textbf{Acknowledgements.}}
We gratefully acknowledge support from the hessian.AI Service Center (funded by the Federal Ministry of Research, Technology and Space, BMFTR, grant no. 16IS22091) and the hessian.AI Innovation Lab (funded by the Hessian Ministry for Digital Strategy and Innovation, grant no. S-DIW04/0013/003). We especially thank Patrick Blauth at the hessian.AI Service Center for technical assistance. This work was further supported by the German Federal Ministry of Education and Research (BMBF): T\"ubingen AI Center, FKZ: 01IS18039A. Gerard Pons-Moll is a member of the Machine Learning Cluster of Excellence, EXC number 2064/1 -- Project number 390727645. The research reported in this publication was also supported by funding from King Abdullah University of Science and Technology (KAUST) -- Center of Excellence for Generative AI, under award number 5940 and a gift from Google.
\bibliographystyle{splncs04}
\bibliography{main}
\end{document}